\documentclass[]{fairmeta}

\usepackage[utf8]{inputenc} 
\usepackage[T1]{fontenc}    
\usepackage{graphicx}
\usepackage{amsmath}
\usepackage{amssymb}
\usepackage{url}            
\usepackage{booktabs}       
\usepackage{amsfonts}       
\usepackage{nicefrac}       
\usepackage{microtype}      
\usepackage{xcolor}         
\usepackage{multirow}
\usepackage{comment}
\usepackage{colortbl}
\usepackage{tocloft}
\usepackage[ruled,vlined]{algorithm2e}
\usepackage{wrapfig}
\usepackage{tabularx}
\usepackage{textcomp}
\usepackage{stfloats}
\usepackage{verbatim}
\usepackage{titlesec}
\usepackage{adjustbox}
\usepackage{pifont}
\usepackage{tikz}
\usepackage{color}
\usepackage{subcaption}
\usepackage{threeparttable}

\RequirePackage{xspace}
\makeatletter
\DeclareRobustCommand\onedot{\futurelet\@let@token\@onedot}
\def\@onedot{\ifx\@let@token.\else.\null\fi\xspace}

\makeatother

\definecolor{lightblue}{rgb}{0.66, 0.85, 0.95}
\definecolor{blue}{RGB}{0, 0, 255}
\hypersetup{
    colorlinks=true,
    linkcolor=red,
    citecolor=blue,
}
\definecolor{c2}{HTML}{FBD9BD}
\definecolor{c3}{HTML}{fe793d}
\definecolor{c4}{HTML}{eedeb0}
\definecolor{rouse}{rgb}{0.981,0.961,0.941}
\usepackage[most]{tcolorbox}
\tcbset{colback=blue!5!white, colframe=blue!20!white, boxrule=0pt, arc=2mm, left=1mm, right=1mm, top=1mm, bottom=1mm}

\definecolor{adptorange}{RGB}{248, 205, 172}
\definecolor{cmpblue}{RGB}{189, 215, 238}
\definecolor{cmpblue}{RGB}{189, 215, 238}

\definecolor{our_red}{RGB}{232,157,160}
\definecolor{our_blue}{RGB}{136,206,230}
\definecolor{our_orange}{RGB}{246,200,168}
\definecolor{our_green}{RGB}{178,211,164}

\definecolor{attn_code0}{RGB}{247,215,200}
\definecolor{attn_code1}{RGB}{238,169,139}
\definecolor{mlp_code0}{RGB}{204,201,221}
\definecolor{mlp_code1}{RGB}{102,95,153}

\definecolor{token_blue}{RGB}{84, 120, 140}

\usepackage{pifont}       
\usepackage{bbding}       
\usepackage{fontawesome}
\usepackage{xspace}

\usepackage{float}

\newlength\savewidth

\newcolumntype{x}[1]{>{\centering\arraybackslash}p{#1pt}}
\newcolumntype{y}[1]{>{\raggedright\arraybackslash}p{#1pt}}
\newcolumntype{z}[1]{>{\raggedleft\arraybackslash}p{#1pt}}

\renewcommand{\paragraph}[1]{\vspace{1mm}\noindent\textbf{#1}}

\usepackage{colortbl}
\usepackage{xcolor}
\usepackage{wrapfig}

\renewcommand{\paragraph}[1]{\vspace{1.25mm}\noindent\textbf{#1}}

\usepackage{listings}

\definecolor{codeblue}{rgb}{0.25, 0.5, 0.5}
\definecolor{codekw}{rgb}{0.35, 0.35, 0.75}
\lstdefinestyle{Pytorch}{
    language = Python,
    backgroundcolor = \color{white},
    basicstyle = \fontsize{9pt}{8pt}\selectfont\ttfamily\bfseries,
    columns = fullflexible,
    aboveskip=1pt,
    belowskip=1pt,
    breaklines = true,
    captionpos = b,
    commentstyle = \color{codeblue},
    keywordstyle = \color{codekw},
}

\definecolor{green}{HTML}{009000}
\definecolor{red}{HTML}{ea4335}
\definecolor{mygold}{HTML}{feecb3}

\title{Does DINOv3 Set a New Medical Vision Standard?\\
\large Benchmarking 2D and 3D Classification, Segmentation, and Registration}
\author[* \dagger 1]{Che Liu}
\author[* 2]{Yinda Chen}
\author[* 2]{Haoyuan Shi}
\author[* 2]{Jinpeng Lu}
\author[* 7]{Bailiang Jian}
\author[* 5, 7]{Jiazhen Pan}
\author[* 3]{Linghan Cai}
\author[* 4]{Jiayi Wang}
\author[* 10]{Jieming Yu}
\author[* 12]{Ziqi Gao}
\author[* 12]{Xiaoran Zhang}
\author[11]{Long Bai}
\author[5, 7]{Yundi Zhang}
\author[7, 8]{Jun Li}
\author[7, 9]{Cosmin I. Bercea}
\author[5]{Cheng Ouyang}
\author[6]{Chen Chen}
\author[2]{Zhiwei Xiong}
\author[7, 8]{Benedikt Wiestler}
\author[7, 8]{Christian Wachinger}
\author[12]{James S. Duncan}
\author[1, 7, 8]{Daniel Rueckert}
\author[1]{Wenjia Bai}
\author[1]{Rossella Arcucci}

\affiliation[1]{Imperial College London}
\affiliation[2]{University of Science and Technology of China\\}
\affiliation[3]{Dresden University of Technology}
\affiliation[4]{University of Erlangen-Nuremberg}
\affiliation[5]{University of Oxford}
\affiliation[6]{University of Sheffield}
\affiliation[7]{Technical University of Munich (TUM)}
\affiliation[8]{Munich Center for Machine Learning\\}
\affiliation[9]{Helmholtz AI and Helmholtz Munich}
\affiliation[10]{The Hong Kong University of Science and Technology\\}
\affiliation[11]{The Chinese University of Hong Kong}
\affiliation[12]{Yale University}

\contribution[*]{Equal contribution}
\contribution[\dagger]{Corresponding author}

\abstract{
The advent of large-scale vision foundation models, pre-trained on diverse natural images, has marked a paradigm shift in computer vision. However, how the frontier vision foundation models' efficacies transfer to specialised domains such as medical imaging remains an open question. This report investigates whether DINOv3, a state-of-the-art self-supervised vision transformer (ViT) pre-trained on natural images, can directly serve as a powerful, unified encoder for medical vision tasks without domain-specific fine-tuning. To answer this, we benchmark DINOv3 across common medical vision tasks, including 2D and 3D classification, segmentation, and registration on a wide range of medical imaging modalities. We systematically analyse its scalability by varying model sizes and input image resolutions. Our findings reveal that DINOv3 shows impressive performance and establishes a formidable new baseline. Remarkably, it can even outperform medical-specific foundation models like BiomedCLIP and CT-Net on several tasks, despite being trained solely on natural images. However, we identify clear limitations: The model's features degrade in scenarios requiring deep domain specialisation, such as in whole-slide images (WSIs), electron microscopy (EM), and positron emission tomography (PET). Furthermore, we observe that DINOv3 does not consistently follow the scaling law in the medical domain. Its performance does not reliably increase with larger models or finer feature resolutions, showing diverse scaling behaviours across tasks. Overall, our work establishes DINOv3 as a strong baseline, whose powerful visual features can serve as a robust prior for multiple medical tasks. This opens promising future directions, such as leveraging its features to enforce multiview consistency in 3D reconstruction.
}

\date{\today}
\correspondence{\texttt{che.liu21@imperial.ac.uk}}

\begin{document}
\thispagestyle{firstheader}
\maketitle
\pagestyle{empty}

\section{Motivation} \label{sec:introduction}
Foundation models, exemplified by Large Language Models (LLMs) \cite{OpenAI2022ChatGPT}, have demonstrated that immense knowledge can be learned from vast, unannotated corpora through self-supervised objectives, leading to impressive scaling laws \cite{kaplan2020scaling}. While this principle has been extended to and often assumed in computer vision, a definitive answer on scaling laws for visual pre-training has been more elusive \cite{alabdulmohsin2022revisiting,xie2023data,el2024scalable}. Recent works have questioned traditional scaling limits, but their evaluation was often focused on narrower tasks \cite{pan2025beyond, pan2025medvlm,fan2025scaling}, leaving their general-purpose capabilities less explored. The DINO series \cite{dinov2,caron2021emerging,dinov3}, in contrast, has been instrumental in showing that self-supervised learning (SSL) can produce emergent visual representations of remarkable quality. Most recently, DINOv3 \cite{dinov3} has pushed this frontier by scaling the visual encoder up to a 7B parameter scale on 1.7B images, demonstrating unprecedented generalization and strong performance across a wide range of visual tasks.

This progress in the natural image domain is highly relevant to medical image analysis, a field that strongly relies on the quality of visual representations to capture subtle anomalies. Indeed, very recent work \cite{yang2025segdino, li2025meddinov3} has shown promising performance using DINOv3 features on specific medical tasks, although the results often depend on careful hyperparameter tuning, leaving the broader impact less clear. The medical domain is characterized by a vast diversity of imaging modalities, from 2D grayscale X-rays \cite{Wang2017ChestXRay8HC} to multi-channel RGB histopathology \cite{CONCH} and 3D volumetric scans \cite{ct-rate}, each demanding distinct visual understanding capabilities. This is further complicated with long-tailed distributions over conditions and the prohibitive cost and regulatory concerns associated with data collection. This heterogeneity and data scarcity highlight the imperative need for strong vision representation extractors. However, the development of a large-scale medical visual foundation model has been hampered by the relative scarcity of curated data due to cost, privacy, and regulatory concerns. Existing efforts, such as BiomedCLIP \cite{BiomedCLIP}, have attempted to bridge this gap by training visual encoders on web-crawled medical images from research articles with text supervision. While valuable, this approach is limited by the quality and scalability of its data source and still relies on language supervision. This dichotomy leads us to a series of fundamental questions: 
\begin{tcolorbox}[colback=mygold, colframe=mygold!75!black, boxrule=0pt, arc=2mm, left=1mm, right=1mm, top=1mm, bottom=1mm]
\begin{center}
\textit{\textbf{Q1: Can DINOv3's \cite{dinov3} natural-image representations excel on medical vision tasks?}}
\end{center}
\end{tcolorbox}

\begin{tcolorbox}[colback=mygold, colframe=mygold!75!black, boxrule=0pt, arc=2mm, left=1mm, right=1mm, top=1mm, bottom=1mm]
\begin{center}
\textit{\textbf{Q2: Does scaling visual pre-training on natural images improve performance in the medical domain?}}
\end{center}
\end{tcolorbox}

\begin{tcolorbox}[colback=mygold, colframe=mygold!75!black, boxrule=0pt, arc=2mm, left=1mm, right=1mm, top=1mm, bottom=1mm]
\begin{center}
\textit{\textbf{Q3: Are the benefits of scaling model size and dataset size transferable across diverse medical tasks and modalities?
}}
\end{center}
\end{tcolorbox}

\section{Benchmark Setup}
To evaluate the capabilities of DINOv3 \cite{dinov3} as a generic off-the-shelf vision encoder for medical imaging, we designed a multi-faceted benchmark assessing its performance across the most common tasks and diverse data formats, ranging from static 2D images and 3D volumetric scans. A key feature claimed in DINOv3 is the fine granularity of its features. Therefore, we are particularly interested in evaluating how this transfers to fine-grained medical imaging tasks such as image segmentation. Our benchmark is structured to cover a wide range of modalities and tasks, including 2D classification, 2D registration, 2D segmentation, 3D classification, 3D segmentation, and 3D registration. The evaluation spans diverse modalities such as X-ray, ultrasound, Whole Slide Imaging (WSI), endoscopy images, Electron Microscopy (EM), and volumetric data from Computed Tomography (CT), Magnetic Resonance Imaging (MRI), and Positron Emission Tomography (PET). We systematically analyse scalability by evaluating three different model scales (DINOv3-S, DINOv3-B, and DINOv3-L) across multiple input resolutions.

\subsection{Classification on 2D Medical Images}
Image classification is a foundational task in medical imaging, often used for diagnostic purposes on planar images or individual video frames. For these tasks, we process 2D images directly as input for the DINOv3 encoder. To accommodate DINOv3's 3-channel input requirement, single-channel grayscale images are replicated three times to create a 3-channel tensor. For native RGB images, such as those from Whole Slide Imaging (WSI) or endoscopic feeds, we use the original data without modification. We benchmark the 2D classification performance on the following publicly available datasets:

\noindent \textbf{NIH-14} \cite{Wang2017ChestXRay8HC} 
This dataset is a large collection of chest X-ray images for multi-label classification of 14 common thoracic pathologies, comprising 112,120 images from 30,805 unique patients. For our experiments, we adhere strictly to the official patient-wise data splits provided by the dataset creators to ensure reproducibility.

\noindent \textbf{RSNA-Pneumonia} \cite{rsna2018pneumonia} 
This dataset from the RSNA Pneumonia Detection Challenge consists of 29,700 chest X-ray images for pneumonia classification. To ensure a standardized comparison, we follow the data splitting methodology proposed in the MGCA \cite{mgca}, which provides a well-defined protocol for training and testing.

\noindent \textbf{Camelyon16} \cite{Camelyon16} 
This dataset comprises 399 H\&E-stained lymph node WSIs for breast cancer metastasis detection (tumor vs. normal). We adopt a 5-fold cross-validation protocol on the Camelyon16 \cite{Camelyon16} training set and additionally report results under the official split for comparability. Under the official split (270 train / 129 test slides), we train on the Camelyon16 \cite{Camelyon16} training set and report performance on its official test set. To assess cross-cohort generalization, we train models with five different random seeds on Camelyon16 \cite{Camelyon16} and evaluate them on the Camelyon17 \cite{Camelyon17} Unseen subset.

\noindent \textbf{Camelyon17} \cite{Camelyon17} 
This dataset is a multi-center cohort for pathological N-staging. The official training set contains 100 patients with 5 labeled slides per patient. Each slide is annotated as negative, micro-metastasis, macro-metastasis, or isolated tumor cells (ITC). In our evaluation protocol, we use Camelyon17 \cite{Camelyon17} solely as an out-of-distribution testbed for models trained on Camelyon16 \cite{Camelyon16}. Since the official test annotations are unavailable, we evaluate on the official training set. Following prior practice \cite{attrimil}, we remove the ITC slides and split Camelyon17 \cite{Camelyon17} into Seen (140 slides) and Unseen (324 slides) subsets based on center overlap with Camelyon16; we report generalization on the Unseen subset in the WSI tumor detection benchmark.

\noindent \textbf{BCNB} \cite{BCNB} 
This dataset is an Early Breast Cancer Core-Needle Biopsy WSI dataset. It contains  1058 patients with molecular status labels: ER (831 positive / 227 negative), PR (790 positive/ 268 negative), HER2 (277 positive/ 781 negative), and Ki67 (156 positive / 902 negative). WSIs are annotated with tumor type, molecular status, number of lymph node metastases, and axillary lymph node (ALN) metastatic status, among others. Using CLAM \cite{CLAM}, we remove background and crop each slide into 224$\times$224 patches at the native resolution, yielding on average $\sim$968 patches per slide. For the BCNB benchmark, we perform 5-fold cross-validation with a 7:1:2 split ratio (train:val:test) within each fold, and evaluate five tasks: ALN metastatic status (N0 vs. N+), and the molecular status prediction of ER, PR, HER2, and Ki67. Unless otherwise specified, preprocessing and tiling are identical across tasks.

\noindent \textbf{Kvasir-Capsule} \cite{Smedsrud2021} 
This dataset represents the largest publicly available PillCAM dataset, comprising 47,238 labeled frames derived from endoscopic feeds depicting various anatomical landmarks with both normal and pathological features. We utilize the 11 categories containing at least 50 samples each: Angiectasia, Fresh blood, Erosion, Erythema, Foreign body, Ileocecal valve, Lymphangiectasia, Normal clean mucosa, Pylorus, Reduced mucosal view, and Ulcer.

\noindent \textbf{AutoLaparo} \cite{Wang2022} 
This dataset contains 21 videos that include 7 unique surgical phases. Each video is recorded at a resolution of 1920 × 1080 pixels and a frame rate of 25 fps, with an average length of about 66 minutes. The dataset is divided into 10 training, 4 validation, and 7 testing videos. To reduce computational demands and emphasize the central region of the surgical field, all videos undergo preprocessing in which they are downsampled to 1 fps and each frame is resized to 250 × 250 pixels, in line with the original preprocessing setup. This process yields a sequence of approximately 83,160 discrete 2D images representing 7 unique surgical phases

\subsection{Registration on 2D Medical Images}
Medical image registration aligns anatomical structures across different temporal or spatial views to facilitate motion tracking and comparative analysis. We evaluate the capacity of DINOv3 to drive precise deformable registration on 2D cardiac ultrasound sequences.

\noindent\textbf{Ultrasound CAMUS} \cite{leclerc2019camus}
The CAMUS dataset (Cardiac Acquisitions for Multi-structure Ultrasound) consists of 2D echocardiograms from 500 patients. Each subject includes apical two-chamber (2CH) and four-chamber (4CH) views at end-diastole (ED) and end-systole (ES), along with ground truth segmentations for the LV cavity, myocardium, and left atrium. Preprocessing involves resampling images to $128 \times 128$ dimensions and normalizing orientation. For experimentation, sequences are grouped by patient and view to ensure valid ED-ES correspondence, resulting in 800 training, 100 validation, and 100 testing registration pairs. HD95 and ASD are reported in pixels.

\subsection{Segmentation on 2D Medical Images}

We evaluate 2D semantic segmentation performance on individual frames derived from medical video sequences. We benchmark on the following publicly available datasets:

\noindent \textbf{EndoVis 2018} \cite{EndoVis2018} 
Originating from the 2018 Robotic Scene Segmentation Challenge, this dataset consists of high-resolution (1280$\times$1024) surgical frames recorded by the da Vinci Xi system during robotic procedures. We adhere to the standard evaluation protocol and data split defined in \cite{gonzalez2020isinet}, utilizing frames from 11 sequences for training and 4 for testing. The task involves pixel-wise segmentation of seven distinct instrument categories: bipolar forceps, prograsp forceps, large needle driver, monopolar curved scissors, ultrasound probe, suction instrument, and clip applier.

\noindent \textbf{EDD 2020} \cite{https://doi.org/10.21227/f8xg-wb80} 
This dataset was established for the 2020 Endoscopy Disease Detection and Segmentation Challenge. It comprises 380 annotated frames collected from multiple international centers, capturing various gastrointestinal organs (colon, esophagus, and stomach) across different endoscopic modalities. We treat this as a multi-class 2D segmentation task targeting five specific pathologies: Barrett’s Oesophagus, suspicious regions, high-grade dysplasia, cancer, and polyp.

\subsection{Classification on 3D Medical Images}
To perform 3D classification with a 2D-native encoder like DINOv3, we adopt a slice-wise feature extraction strategy. We process each 2D slice of a 3D volume independently through the DINOv3 backbone to obtain a feature embedding for that slice. The resulting set of slice embeddings is then aggregated into a single feature vector representing the entire volume, typically via mean pooling \cite{muller2025medical}. As with the 2D tasks, grayscale slices are replicated across three channels before being fed into the model. For this task, the model's performance is assessed using the following publicly available dataset:

\noindent\textbf{CT-RATE \cite{ct-rate}} 
This dataset is a large-scale collection of 3D medical imaging, pairing 47k non-contrast CT volumes(20k patients) with their corresponding radiology reports. The dataset is annotated for 18 clinically significant abnormalities. For all of our experiments, we utilize the official data splits provided by the organizers for training and evaluation procedures, extracted features from every slice of these over 40,000 volumes and employed two methods for the downstream classification task: zero-shot k-nearest neighbors (k-NN) and linear probing. In the CT-RATE original work \cite{ct-rate}, the associated dataset is annotated with multi-label binary labels. This annotation scheme specifies for each clinical category whether a case has a particular condition or does not have that condition. Consequently, this task can be viewed as a multi-label binary classification problem, where normal/abnormal binary classifications are performed across multiple categories.

\subsection{Segmentation on 3D Medical Images}
Segmentation on 3D medical images is the task of producing a dense, voxel-wise prediction to delineate anatomical structures or pathologies within a volumetric scan. To achieve this with a 2D encoder, we process the volume on a slice-by-slice basis. The 2D feature map extracted from each slice by the DINOv3 encoder is preserved. These 2D feature maps are then stacked to construct a pseudo-3D feature volume, which serves as the input to a lightweight segmentation head that produces the final voxel-wise predictions. In our evaluation, we freeze the vision encoder and only fine-tune the segmentation head. We benchmark this task on 14 widely-used public datasets:

\noindent\textbf{Medical Segmentation Decathlon (MSD) \cite{msd}} The MSD challenge provides 10 distinct 3D medical image segmentation tasks across various modalities and body parts. Since the official online evaluation platform is no longer available, we adopt a 5-fold cross-validation approach on the public training set. Following the standard protocol established in previous medical SSL works~\cite{wu2024voco}, we normalize all volumes and apply standard geometric augmentations, including random rotations and flips. For each fold, we use a random 80\%/20\% split for training and validation, reporting the average performance across all folds.

\noindent\textbf{EM Neuron Segmentation in CREMI \cite{CREMI2016}} The CREMI dataset originates from the 2016 CREMI challenge, designed to advance neuron segmentation in electron microscopy volumes. The data are from an adult Drosophila brain imaged at a resolution of 4 × 4 × 40 nm with 1250 × 1250 pixels per slice. It includes three subsets, CREMI-A, CREMI-B, and CREMI-C, each providing 125 annotated slices that represent different neuron types. The difficulty increases from A to C, with later subsets exhibiting more intricate neuronal morphology. In our setup, we train on the first 100 sections from each subset and evaluate on the remaining 25 sections.

\noindent\textbf{EM Neuron Segmentation in AC3/4 \cite{AC34}} Both AC3 and AC4 are densely annotated EM volumes from the Kasthuri15 dataset \cite{AC34}, acquired at 3 × 3 × 29 nm resolution with 1024 × 1024 pixels per slice. AC3 comprises 256 consecutive sections and exhibits greater structural heterogeneity, leading to higher topological complexity. AC4 contains 100 sections with relatively uniform contrast, providing a stable target for optimization. In our experiments, we train on the first 80 sections of AC4 and evaluate on the first 100 sections of AC3.

\noindent\textbf{Automated Lesion Segmentation in Whole-Body FDG-PET/CT Challenge (AutoPET-II) \cite{autopetii}} The autoPET-II challenge provides a comprehensive dataset of 1014 whole-body FDG-PET/CT scans for automated tumor lesion segmentation in oncology. The dataset focuses on malignant melanoma, lymphoma, and lung cancer lesions across diverse patient populations. Following established evaluation protocols, we utilize the official train/validation split provided by the organizers. All volumes are preprocessed with intensity normalization, and we apply standard data augmentation techniques including random rotations and flips to enhance model robustness.

\noindent\textbf{Head and Neck Tumor segmentation and outcome prediction in PET/CT images (HECKTOR 2022) \cite{hecktor2022}} The HECKTOR 2022 dataset comprises 882 head and neck FDG-PET/CT scans with annotations for primary gross tumor volume (GTVp) and lymph node gross tumor volume (GTVn). This dataset presents unique challenges due to the complex anatomy of the head and neck region and the heterogeneous appearance of head and neck cancers. We follow the challenge's standard preprocessing pipeline, which includes image registration between PET and CT modalities and intensity normalization. The evaluation follows the official challenge protocol to ensure fair comparison with published benchmarks.

\subsection{Registration on 3D Medical Images}

\noindent\textbf{MRI ACDC} \cite{ACDC}
We utilize the ACDC dataset, comprising cardiac MRI volume sequences from 150 patients. Each sequence includes frames at end-diastole (ED) and end-systole (ES), along with corresponding segmentation maps for the LV cavity, myocardium, and right ventricle. Preprocessing includes resampling to a uniform $(1.5, 1.5, 3.15)$ mm spacing and myocardium-centered cropping to volume dimensions of $128 \times 128 \times 32$. Intensities are linearly normalized to $[-1, 1]$. For experimentation, the dataset is split into 80 training, 20 validation, and 50 testing patients. HD95 and ASD are reported in millimeters, accounting for volume anisotropy.

\begin{table*}[t!]
\centering
\caption{Overview of datasets included in the DINOv3 medical imaging benchmark, spanning 2D and 3D modalities across classification, segmentation, and registration.}
\label{tab:dataset_overview}

\begin{tabularx}{\textwidth}{@{}c c c >{\centering\arraybackslash}X@{}}
\toprule
\textbf{Dataset} & \textbf{Modality} & \textbf{Data Scale} & \textbf{Target} \\
\midrule

\multicolumn{4}{c}{\textit{\textbf{2D Classification}}} \\ 
NIH-14 & Chest X-ray & 112,120 images & 14 thoracic pathologies \\
RSNA-Pneumonia & Chest X-ray & 29,700 images & Pneumonia detection \\
Camelyon16 & WSI (H\&E) & 399 slides & Tumor metastasis detection \\
Camelyon17 & WSI (H\&E) & 500 slides & Pathological N-staging \\
BCNB & WSI (Biopsy) & 1,058 patients & Molecular \& ALN status prediction \\
Kvasir-Capsule & Capsule Endoscopy & 47,238 frames & 11 anatomical/pathological classes \\
AutoLaparo & Laparoscopy & 83,160 frames & 7 surgical phase recognition \\\midrule

\multicolumn{4}{c}{\textit{\textbf{2D Segmentation}}} \\ 
EndoVis 2018 & Robotic Surgery & 15 sequences & 7 surgical instruments segmentation \\
EDD 2020 & Endoscopy & 380 frames & 5 disease classes segmentation \\
\midrule

\multicolumn{4}{c}{\textit{\textbf{3D Classification}}} \\ 
CT-RATE & CT & 47,000 volumes & 18 clinical abnormalities classification \\
\midrule

\multicolumn{4}{c}{\textit{\textbf{3D Segmentation}}} \\ 
MSD & CT / MRI & 10 tasks & 10 organ/tumor targets \\
CREMI (A/B/C) & EM (Drosophila) & $3 \times 125$ slices & Neuron segmentation \\
AC3/4 & EM (Mouse) & 356 sections & Neuron segmentation \\
AutoPET-II & FDG-PET/CT & 1,014 scans & Whole-body lesion segmentation \\
HECKTOR 2022 & FDG-PET/CT & 882 scans & Head \& Neck tumor segmentation \\
\midrule

\multicolumn{4}{c}{\textit{\textbf{Registration}}} \\ 
CAMUS & Ultrasound & 1,000 pairs & 2D Cardiac ED-ES registration \\
ACDC & Cardiac MRI & 150 patients & 3D Cardiac ED-ES registration \\

\bottomrule
\end{tabularx}
\end{table*}

\section{Task Adaptation}
To assess the quality of the visual features produced by DINOv3 \cite{dinov3}, we apply straightforward, standardized adaptation techniques that introduce minimal task-specific parameters. This design ensures the benchmark primarily reflects the strength of the frozen representations.

\subsection{Classification}
Our primary evaluation protocol for the 2D X-ray, 2D endoscopic, and 3D CT datasets is linear probing. In this setting, the DINOv3 \cite{dinov3} backbone remains frozen, and only a single linear layer is trained on top of the extracted features using binary cross-entropy (BCE) loss with a learning rate of 0.005, a batch size of 1024, and for 50 epochs.

For the CT-RATE \cite{ct-rate} dataset, we additionally perform k-nearest neighbors (k-NN) evaluation. We extract feature embeddings for all scans, and for each of the 18 disease categories (treated as independent binary tasks), k-NN predicts the presence or absence of the disease based on feature similarity.

For whole-slide pathological classification tasks, we use the multiple instance learning (MIL) paradigm. Each WSI is tiled into non-overlapping 224$\times$224 patches and treated as a bag $X=\{\mathbf{x}_i\}_{i=1}^{N}$. Per-patch features are extracted with a frozen DINOv3 encoder (with global average pooling) to obtain $\mathbf{e}_i\in\mathbb{R}^{D_0 \times 1}$. We then apply a learnable linear projection:
\begin{equation}
\mathbf{h}_i=\mathbf{W}_{\mathrm{proj}}\mathbf{e}_i+\mathbf{b}_{\mathrm{proj}}, 
\qquad \mathbf{W}_{\mathrm{proj}}\in\mathbb{R}^{D\times D_0},\ \mathbf{b}_{\mathrm{proj}}\in\mathbb{R}^{D\times 1},\ \mathbf{h}_i\in\mathbb{R}^{D\times 1},\ D<D_0.
\end{equation}
Instance embeddings are aggregated using attention-based deep multiple instance learning (ABMIL) \cite{ilse2018attention}:
\begin{equation}
a_i = 
\frac{\exp\!\big\{\mathbf{w}^\top\!\big(\tanh(\mathbf{V}\mathbf{h}_i)\odot \sigma(\mathbf{U}\mathbf{h}_i)\big)\big\}}
{\displaystyle \sum_{j=1}^{N} \exp\!\big\{\mathbf{w}^\top\!\big(\tanh(\mathbf{V}\mathbf{h}_j)\odot \sigma(\mathbf{U}\mathbf{h}_j)\big)\big\}},
\qquad
\mathbf{z}=\sum_{i=1}^{N} a_i \mathbf{h}_i,
\end{equation}
where $\mathbf{U},\mathbf{V}\in\mathbb{R}^{H\times D}$, $\mathbf{w}\in\mathbb{R}^{H \times 1}$, $\sigma(\cdot)$ denotes the sigmoid function, $a_i\in\mathbb{R}$ and $\sum_i a_i=1$, and $\mathbf{z}\in\mathbb{R}^{D \times 1}$ is the slide-level representation. A task-specific head $g(\cdot)$ maps $\mathbf{z}$ to $\hat{Y}$; training uses bag-level cross-entropy. Unless specified, the DINOv3 encoder is frozen and only the projection, attention, and head layers are trained.

\subsection{Segmentation}
The 2D segmentation architecture consists of three main components: (1) a frozen DINOv3 encoder that extracts dense features from images of arbitrary size; (2) a lightweight 2D adaptive decoder that refines and progressively upsamples the feature maps; and (3) a segmentation head that produces pixel-wise logits. The decoder employs shallow convolutional blocks with dynamic bilinear upsampling to align predictions with the target resolution during training and inference.

For 3D medical image segmentation, we leverage DINOv3's 2D feature extraction capabilities in a slice-wise manner. Each axial slice of the 3D volume is processed independently through the frozen DINOv3 encoder to extract dense feature maps. These 2D feature maps are then stacked along the slice dimension to construct a pseudo-3D feature volume.

The segmentation architecture consists of three main components: (1) the frozen DINOv3 encoder for feature extraction, (2) a lightweight 3D decoder that processes the pseudo-3D features, and (3) a segmentation head that produces voxel-wise predictions. The decoder employs 3D convolutional layers with skip connections to progressively upsample features to the original volume resolution.

For the MSD benchmark, we adopt the established 5-fold cross-validation protocol to ensure robust evaluation. Each fold uses an 80\%/20\% split for training and validation, with careful attention to maintaining patient-level separation to avoid data leakage. All models are trained using the Dice loss function combined with cross-entropy loss, optimized with AdamW optimizer using a learning rate of 1e-4 and a cosine annealing schedule.

For the AutoPET-II and HECKTOR 2022 benchmarks, we followed the official challenge protocol, using an 80\%/20\% split for training and validation to maintain consistency with the published benchmarks. Models were trained using a combination of the Dice and CE losses and optimized with the AdamW optimizer, a learning rate of 1e-4, and a linear warmup and cosine annealing schedule.

For the EM neuron segmentation benchmarks, CREMI and AC3/4, we follow the experimental protocols established in previous studies to ensure direct comparability. Models are trained using a weighted mean squared error objective and optimized with the Adam optimizer, a learning rate of 1e-3. During inference, instance segmentations are obtained using the Waterz \cite{MALA} post-processing method.

\subsection{Registration}
To adapt the 2D DINOv3 backbone for 3D volumetric data, a slice-wise feature extraction strategy is employed, mirroring the approach used for segmentation models. Each axial slice of the 3D volume is independently passed through the frozen DINOv3 encoder to obtain dense 2D feature maps. These maps are subsequently stacked along the slice dimension to form a pseudo-3D feature volume. Following the DINO-Reg methodology \cite{song2024dino}, the high-dimensional features extracted from both the fixed and moving volumes are aggregated, and Principal Component Analysis (PCA) is applied to learn a shared, compressed basis. These low-dimensional PCA features serve as the input to the self-supervised registration network. This network utilizes a lightweight, 3D U-Net-like structure which accepts the feature volumes. The network is trained to predict a dense, 3D deformation displacement field, which is then applied to the moving image to generate the registered output. For 2D ultrasound registration, an analogous 2D U-Net architecture processes the 2D feature maps directly.

\subsection{Evaluation Metrics}

\noindent\textbf{Classification:} We report the Area Under the Curve (AUC), accuracy, precision, recall, and F1-score. For multi-label tasks such as NIH-14 \cite{Wang2017ChestXRay8HC} and CT-RATE \cite{ct-rate}, these metrics are averaged across classes. For endoscopic datasets, we additionally report the Jaccard index (for surgical phase recognition) as well as both macro-averaged and weighted-averaged precision, recall, and F1-scores to account for class imbalance.

\noindent\textbf{Segmentation:} For 3D segmentation tasks, we report the mean Dice score for the MSD \cite{msd} datasets. For the PET datasets, we report the Dice score, HD95, precision, and recall. for the EM datasets, we use the Variation of Information (VOI)~\cite{voi} and Adapted Rand Error (ARAND) \cite{arand}.

\noindent\textbf{Registration:} For single-modality registration tasks, we evaluate our results quantitatively by warping
 segmentation maps in the source image with our predicted displacement and compute anatomical conformance in terms of Dice Similarity Coefficient (DSC), 95th percentile Hausdorff Distance (HD95), and Average Surface Distance (ASD). HD95 and ASD are calculated using medpy.metric.binary implementations.

\section{Experiments}
\subsection{2D Classification Results}
\begin{figure}[ht!]
    \centering
    \begin{subfigure}{\linewidth}
        \centering
        \includegraphics[width=\linewidth]{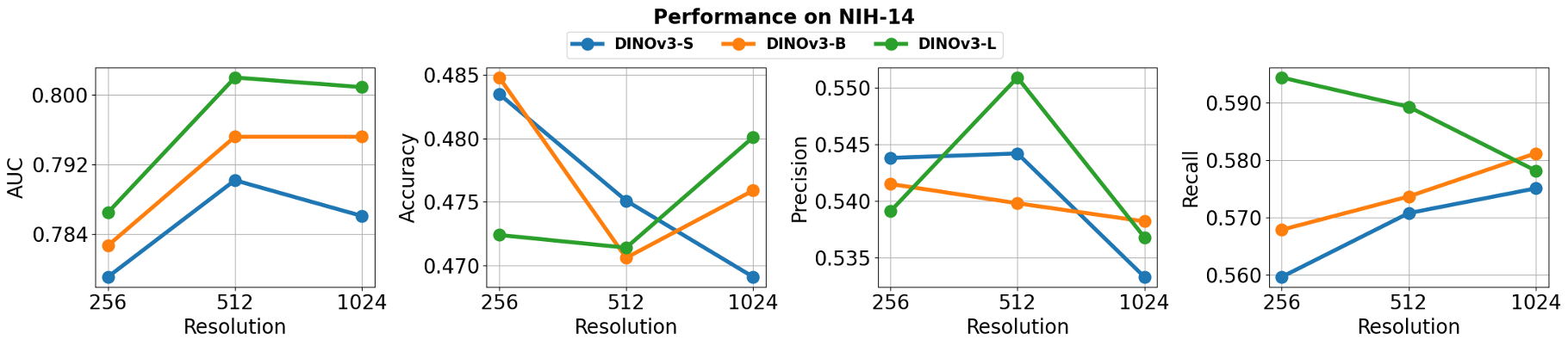}
        \caption{NIH-14 dataset.}
        \label{fig:nih-scale}
    \end{subfigure}

    \begin{subfigure}{\linewidth}
        \centering
        \includegraphics[width=\linewidth]{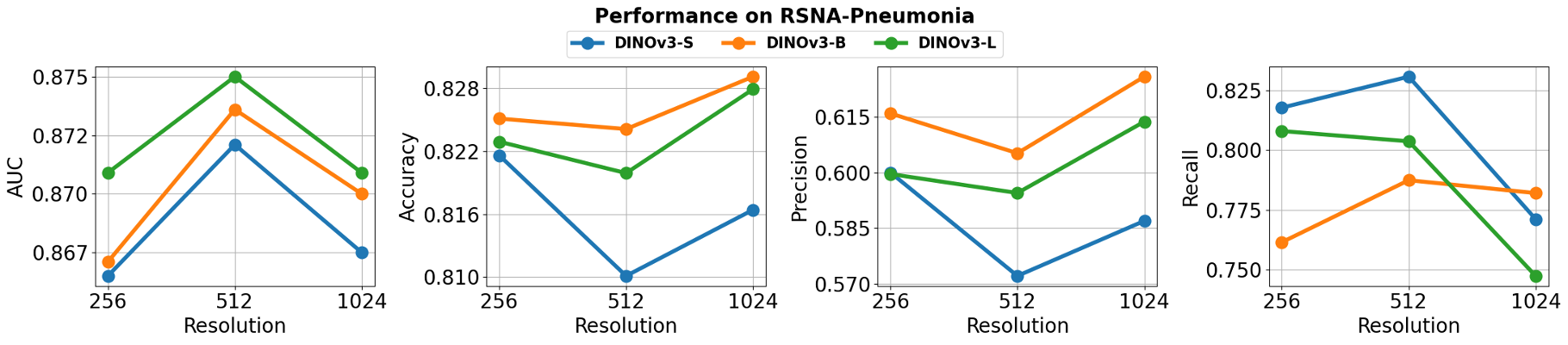}
        \caption{RSNA-Pneumonia dataset.}
        \label{fig:rsna-scale}
    \end{subfigure}

    \caption{Scaling behavior of DINOv3 models across datasets. The results reveal a non-trivial relationship between performance, model size, and input resolution, where larger models or higher resolutions do not consistently yield better outcomes.}
    \label{fig:scaling}
\end{figure}

\noindent\textbf{Classification on Chest X-ray images:} 
\label{exp:xray}
On the NIH-14 and RSNA-Pneumonia chest X-ray datasets, DINOv3 models demonstrate strong, competitive performance. As shown in Table \ref{tab:xray}, DINOv3-L achieves the highest AUC on NIH-14, outperforming the medical-specific BiomedCLIP model. While BiomedCLIP performs best on the RSNA-Pneumonia task, DINOv3 models are close contenders. However, the results also highlight an inconsistent scaling behavior, as seen in Figure \ref{fig:scaling}. Performance does not reliably improve with larger model sizes or higher input resolutions; for instance, AUC for all models on NIH-14 peaks at a 512x512 resolution before declining. This suggests that simply increasing model scale does not guarantee better performance in this domain.

\begin{table}[ht!]
\caption{2D classification linear probing results comparing baseline and DINOv3 series on the NIH-14 and RSNA-Pneumonia datasets. All models use an input resolution of 256x256. For each metric, the highest performing method is marked in \textbf{bold}, and the second highest is \underline{underlined}.}
\centering
\begin{tabular}{lcccccccc}
\toprule
\textbf{Methods} & \multicolumn{4}{c}{\textbf{NIH-14}} & \multicolumn{4}{c}{\textbf{RSNA-Pneumonia}} \\
\cmidrule(lr){2-5} \cmidrule(lr){6-9}
& \textbf{AUC} & \textbf{ACC} & \textbf{Precision} & \textbf{Recall} & \textbf{AUC} & \textbf{ACC} & \textbf{Precision} & \textbf{Recall} \\
\midrule
BiomedCLIP \cite{BiomedCLIP} & 0.7771 & 0.4820 & \textbf{0.5454} & 0.5643 & \textbf{0.8831} & \textbf{0.8374} & \textbf{0.6368} & \underline{0.8026} \\
DINOv3-S \cite{dinov3} & 0.7788 & \textbf{0.4838} & 0.5419 & \textbf{0.5791} & 0.8667 & 0.8221 & 0.6048 & \textbf{0.8156} \\
DINOv3-B \cite{dinov3} & \underline{0.7833} & \underline{0.4823} & \underline{0.5446} & 0.5753 & 0.8666 & \underline{0.8274} & \underline{0.6227} & 0.7679 \\
DINOv3-L \cite{dinov3} & \textbf{0.7865} & 0.4674 & 0.5355 & \underline{0.5779} & \underline{0.8708} & 0.8209 & 0.5972 & 0.7744 \\
\bottomrule
\end{tabular}
\label{tab:xray}
\end{table}

\begin{figure}[ht!]
\centering
\includegraphics[width=1.0\textwidth]{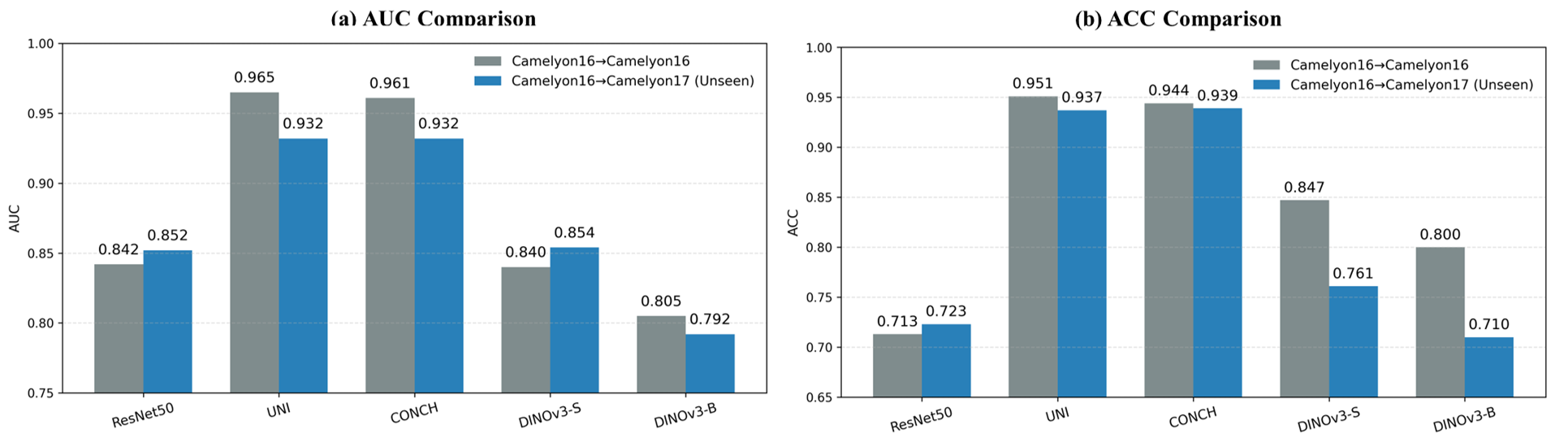}
\caption{Cross-domain generalization on Camelyon16 \cite{Camelyon16} and Camelyon17 \cite{Camelyon17}: In-domain vs. Out-of-domain AUC and ACC comparisons.}
\label{fig:bar}
\end{figure}

\begin{figure}[ht!]
\centering
\includegraphics[width=1.0\textwidth]{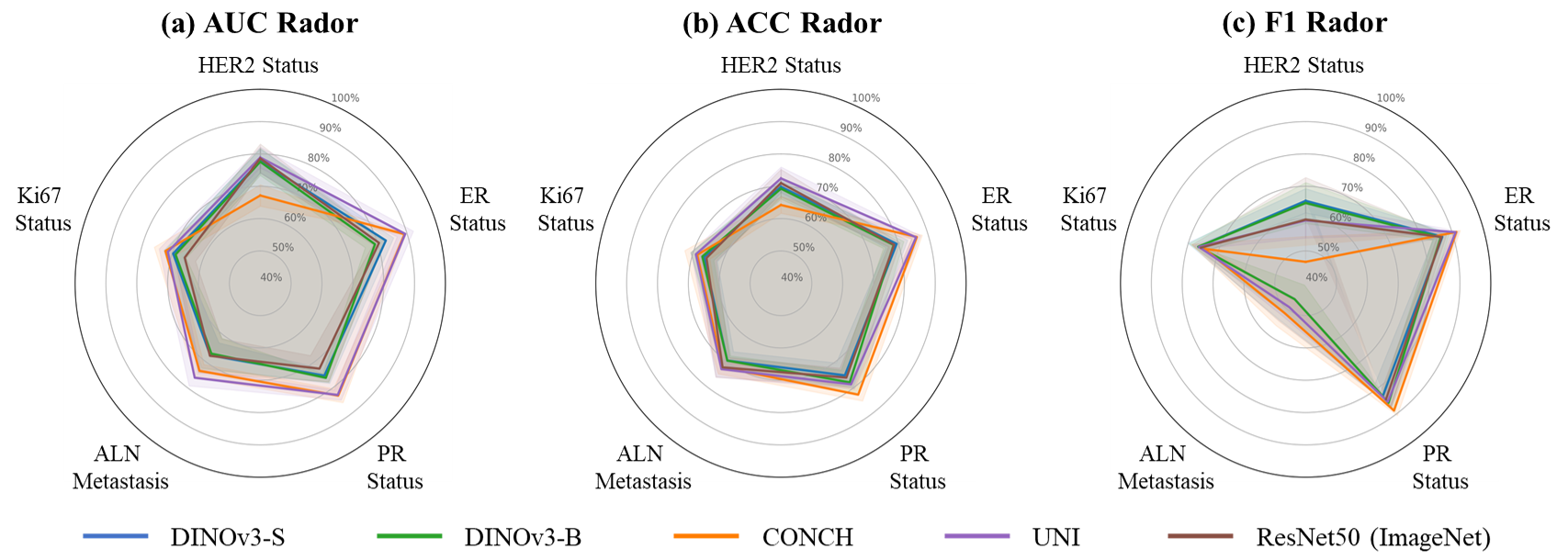}
\caption{Performance comparison across ALN metastasis and receptor status tasks on the BCNB \cite{BCNB} dataset. The default feature aggregator for the whole-slide images is the attention-based multiple instance learning method \cite{ilse2018attention}.}
\label{fig:rador}
\end{figure}

\noindent\textbf{Classification on Pathology images:} 
\label{exp:wsi}
In the domain of WSIs, DINOv3's performance is significantly weaker than specialized models. For both the Camelyon16~\cite{Camelyon16} and Camelyon17~\cite{Camelyon17} datasets, as shown in Tables~\ref{tab:in_domain_c16} and~\ref{tab:ood_c17} and Figure~\ref{fig:bar}, DINOv3 models are substantially outperformed by pathology-specific foundation models like UNI~\cite{UNI} and CONCH~\cite{CONCH}. Their performance is only comparable to a generic ResNet50~\cite{ResNet} baseline, indicating that DINOv3's natural image features do not effectively transfer to the fine-grained, textural analysis required for histopathology. This limitation is further confirmed by the radar charts for the BCNB dataset in Figure~\ref{fig:rador}, where DINOv3 again lags behind the domain-specialized models across multiple molecular subtyping tasks.

\begin{table}[ht!]
\caption{In-domain tumour detection on Camelyon16. Patch features are aggregated with ABMIL. Models are trained on the Camelyon16 training set and evaluated on its test set. The highest results are in \textbf{bold} and the second highest are \underline{underlined}.}
\centering
{%
\begin{tabular}{l|cccc}
\toprule
\multirow{2}{*}{\textbf{Patch Encoder}} & \multicolumn{4}{c}{\textbf{Camelyon16} $\rightarrow$ \textbf{Camelyon16}} \\ \cmidrule{2-5}
& \textbf{AUC} & \textbf{ACC} & \textbf{Precision} & \textbf{Recall} \\ \midrule
ResNet50 (ImageNet) \cite{ResNet} & 0.842 & 0.713 & 0.594 & 0.776 \\
UNI \cite{UNI}                    & \textbf{0.965} & \textbf{0.951} & \textbf{0.959} & \textbf{0.938} \\
CONCH \cite{CONCH}                & \underline{0.961} & \underline{0.944} & \underline{0.956} & \underline{0.928} \\ \midrule
DINOv3-S \cite{dinov3}            & 0.840 & 0.847 & 0.898 & 0.682 \\
DINOv3-B \cite{dinov3}            & 0.805 & 0.800 & 0.834 & 0.629 \\ \bottomrule
\end{tabular}}
\label{tab:in_domain_c16}
\end{table}

\begin{table}[ht!]
\caption{Out-of-domain tumour detection on Camelyon17. Models are trained on Camelyon16 and evaluated on Camelyon17 (Unseen). The highest results are in \textbf{bold} and the second highest are \underline{underlined}.}
\centering
{%
\begin{tabular}{l|cccc}
\toprule
\multirow{2}{*}{\textbf{Patch Encoder}} & \multicolumn{4}{c}{\textbf{Camelyon16} $\rightarrow$ \textbf{Camelyon17 (Unseen)}} \\ \cmidrule{2-5}
& \textbf{AUC} & \textbf{ACC} & \textbf{Precision} & \textbf{Recall} \\ \midrule
ResNet50 (ImageNet) \cite{ResNet} & 0.852 & 0.723 & 0.607 & 0.808 \\
UNI \cite{UNI}                    & \textbf{0.932} & \underline{0.937} & \underline{0.933} & \textbf{0.928} \\
CONCH \cite{CONCH}                & \textbf{0.932} & \textbf{0.939} & \textbf{0.934} & \underline{0.913} \\ \midrule
DINOv3-S \cite{dinov3}            & \underline{0.854} & 0.761 & 0.589 & 0.894 \\
DINOv3-B \cite{dinov3}            & 0.792 & 0.710 & 0.529 & 0.820 \\ \bottomrule
\end{tabular}}
\label{tab:ood_c17}
\end{table}

\noindent\textbf{Classification on Endoscopic Imaging.}
\label{exp:endo}
We evaluate DINOv3 on two endoscopic datasets: Kvasir-Capsule (capsule endoscopy) and AutoLaparo (laparoscopy). As shown in Table \ref{tab:kvasir_results}, DINOv3 provides competitive baselines but does not outperform specialized, fully supervised State-of-the-Art (SOTA) methods like VAPCaps \cite{JOSEPH2025131325} on Kvasir-Capsule. However, on the AutoLaparo surgical phase recognition task (Table \ref{tab:autolaparo_results}), DINOv3-L achieves the highest Precision ($77.83\%$) and Jaccard index ($57.65\%$), outperforming recent methods such as STSANet \cite{Li2026STSANet} in these metrics, though STSANet retains the highest accuracy.
\begin{table}[t!]
    \centering
    \caption{Quantitative comparison of DINOv3 with state-of-the-art methods on the \textbf{Kvasir-Capsule} dataset. Best results are highlighted in \textbf{bold}.}
    \label{tab:kvasir_results}
    \resizebox{\textwidth}{!}{%
    \begin{tabular}{l ccc ccc c}
        \toprule
        \textbf{Method} & \multicolumn{3}{c}{\textbf{Macro Average}} & \multicolumn{3}{c}{\textbf{Weighted Average}} & \textbf{Accuracy} \\
        \cmidrule(lr){2-4} \cmidrule(lr){5-7}
         & Precision & Recall & F-1 Score & Precision & Recall & F-1 Score & ($\uparrow$) \\
        \midrule
        GMSRF net \cite{GMSRFNet} & 0.1568 & 0.1980 & 0.1575 & 0.7431 & 0.6095 & 0.6636 & 0.6090 \\
        ConvMix - 1536/20 \cite{ConvMix153620} & 0.1722 & 0.2275 & 0.1697 & 0.7431 & 0.6021 & 0.6524 & 0.6021 \\
        ConViT-S \cite{ConViTS} & 0.1765 & 0.2182 & 0.1689 & 0.7673 & 0.5610 & 0.6312 & 0.5610 \\
        Swin-S \cite{SwinS} & 0.1538 & 0.2388 & 0.1525 & 0.7390 & 0.5800 & 0.6334 & 0.5800 \\
        FocalConv net \cite{FocalConvNet} & 0.2438 & 0.2745 & 0.2178 & 0.7557 & 0.6373 & 0.6734 & 0.6373 \\
        Vats \textit{et al.} \cite{VatsEtAl} & 0.2489 & 0.2541 & 0.2353 & 0.6838 & 0.6671 & 0.6654 & 0.6671 \\
        API net \cite{APINet} & 0.9509 & 0.9808 & 0.9650 & 0.9879 & 0.9873 & 0.9875 & 0.9873 \\
        VAPCaps \cite{JOSEPH2025131325} & \textbf{0.9778} & \textbf{0.9828} & \textbf{0.9800} & \textbf{0.9927} & \textbf{0.9926} & \textbf{0.9926} & \textbf{0.9926} \\
        \midrule
        DINOv3-S \cite{dinov3} & 0.5810 & 0.4804 & 0.5187 & 0.7679 & 0.7640 & 0.7626 & 0.7640 \\
        DINOv3-B \cite{dinov3} & 0.6221 & 0.5138 & 0.5513 & 0.7878 & 0.7766 & 0.7774 & 0.7766 \\
        DINOv3-L \cite{dinov3} & 0.6000 & 0.4978 & 0.5338 & 0.7797 & 0.7675 & 0.7700 & 0.7675 \\
        \bottomrule
    \end{tabular}%
    }
\end{table}

\begin{table}[t!]
    \centering
    \caption{Quantitative comparison of DINOv3 with state-of-the-art methods on the \textbf{AutoLaparo} dataset. Best results are highlighted in \textbf{bold}.}
    \label{tab:autolaparo_results}
    \begin{tabular}{l cccc}
        \toprule
        \textbf{Method} & \textbf{Accuracy} & \textbf{Precision} & \textbf{Recall} & \textbf{Jaccard} \\
        \midrule
        SV-RCNet \cite{Jin2018SVRCNet}   & 75.60 & 64.00 & 59.70 & 47.20 \\
        TMRNet \cite{Jin2021TMRNet}      & 78.20 & 66.00 & 61.50 & 49.60 \\
        Trans-SVNet \cite{Gao2021}       & 78.30 & 64.20 & 62.10 & 50.70 \\
        LoViT \cite{Liu2025LoViT}        & 77.86 & 71.03 & 64.78 & 52.56 \\
        STSANet \cite{Li2026STSANet}     & \textbf{79.48} & 66.21 & 67.07 & 52.58 \\
        \midrule
        DINOv3-S \cite{dinov3}           & 76.17 & 73.64 & \textbf{72.31} & 57.39 \\
        DINOv3-B \cite{dinov3}           & 73.33 & 75.38 & 71.77 & 56.40 \\
        DINOv3-L \cite{dinov3}           & 77.29 & \textbf{77.83} & 70.91 & \textbf{57.65} \\
        \bottomrule
    \end{tabular}
\end{table}

\subsection{2D Segmentation Results}
\begin{figure}[ht!]
\centering
\includegraphics[width=\textwidth]{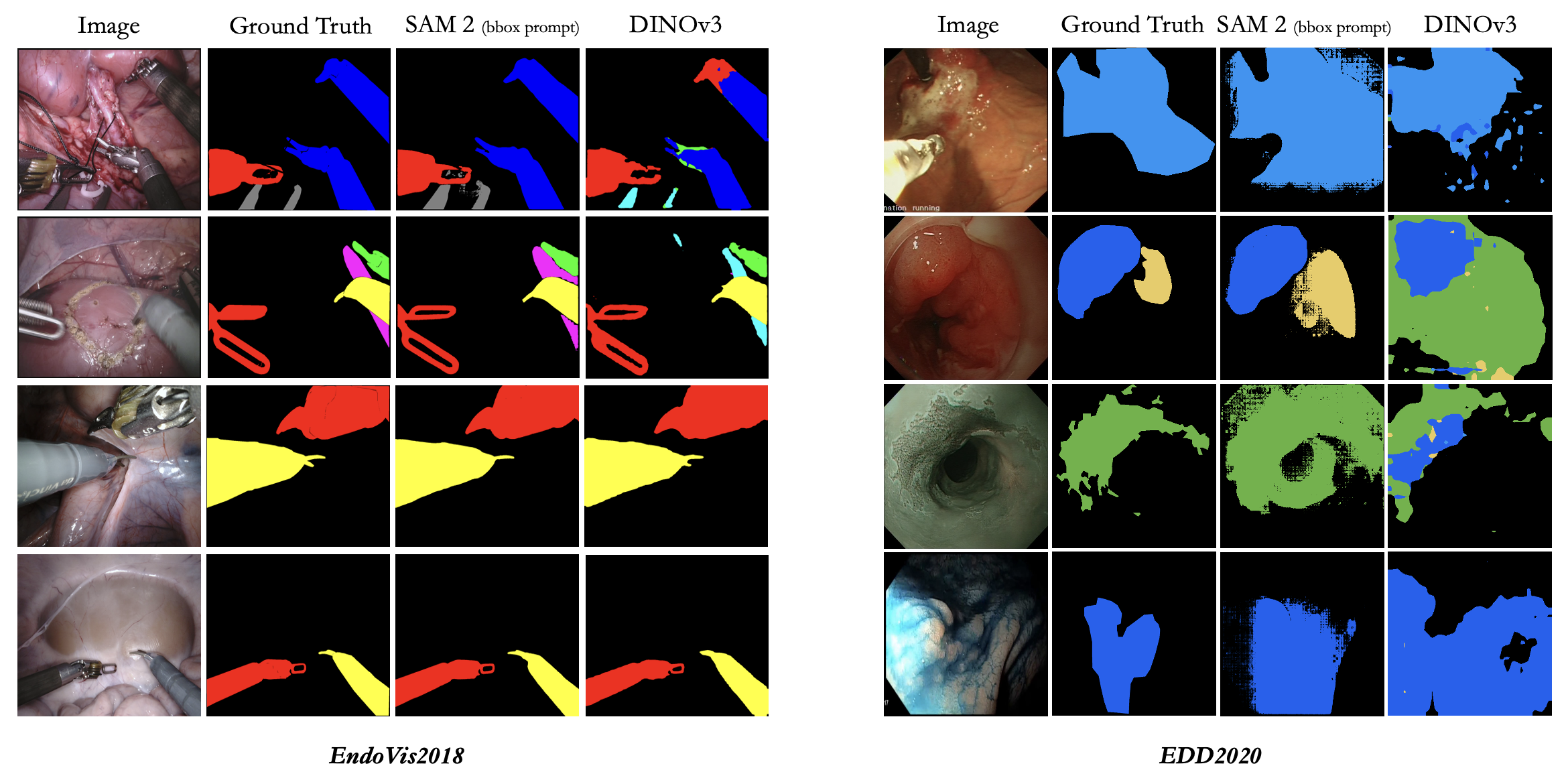}
\caption{Visualization of segmentation results for surgical instruments on the EndoVis18 dataset and disease regions on the EDD2020 dataset.}
\label{fig:endo_edd_visual}
\end{figure}

We present the quantitative results for surgical instrument segmentation on the EndoVis18 \cite{EndoVis2018} dataset in Table \ref{tab:overall_res_endovis18}, where DINOv3-L achieves a state-of-the-art Binary IoU of 92.19\%, surpassing prompt-based methods, although the latter remain superior in fine-grained instrument parsing. This is followed by the disease segmentation results on the EDD 2020 \cite{https://doi.org/10.21227/f8xg-wb80} dataset in Table \ref{tab:edd}, where DINOv3-S achieves a top-ranking Dice score of 93.93\% for polyp segmentation, despite the specialized EAT model yielding higher overall mean IoU. Visualizations of the segmentation performance for both tasks are provided in Figure \ref{fig:endo_edd_visual}.

\begin{table}[htbp]
  \centering
  \caption{Quantitative comparison of DINOv3 with other SOTA methods on the tasks of binary segmentation and instrument segmentation on the EndoVis18 dataset. Results for other SOTA methods are derived from \cite{yu2024sam2roboticsurgery}. Categorical information directly inherits from associated prompts.}
  \begin{threeparttable}
    \resizebox{\columnwidth}{!}{
    \begin{tabular*}{\columnwidth}{@{\extracolsep{\fill}} l l c l c c @{}}
    \toprule
    \multirow{2}[2]{*}{\textbf{Task Type}} & \multirow{2}[2]{*}{\textbf{Methods}} & \multirow{2}[2]{*}{\textbf{Pub/Year(20-)}} & \multirow{2}[2]{*}{\textbf{Arch.}} & \multicolumn{2}{c}{\textbf{EndoVis18}} \\
    \cmidrule{5-6}          &       &       &       & \textbf{Binary IoU} & \textbf{Instrument IoU} \\
    \midrule
    \multirow{5}[2]{*}{Single-Task} 
          & Vanilla UNet \cite{ronneberger2015u} & MICCAI15 & UNet & 68.89 & - \\
          & TernausNet \cite{shvets2018automatic} & ICMLA18 & UNet & - & 46.22 \\
          & MF-TAPNet \cite{jin2019incorporating} & MICCAI19 & UNet & - & 67.87 \\
          & Wang et al. \cite{wang2022rethinking} & MICCAI22 & UNet & 58.12 & - \\
          & ISINet \cite{gonzalez2020isinet} & MICCAI21 & Res50 & - & 73.03 \\
    \midrule
    \multirow{5}[0]{*}{Multi-Task} 
          & ST-MTL \cite{islam2021st} & MedIA21 & - & - & - \\
          & AP-MTL \cite{islam2020ap} & ICRA20 & - & - & - \\
          & S-MTL \cite{seenivasan2022global} & RA-L22 & - & - & 43.54 \\
          & TraSeTR \cite{zhao2022trasetr} & ICRA22 & Res50 + Trfm & - & 76.20 \\
          & S3Net \cite{baby2023forks} & WACV23 & Res50 & - & 75.81 \\
    \midrule
    \multirow{6}[0]{*}{Prompt-based} 
          & SAM (1 Point) \cite{kirillov2023segment} & arxiv23 & ViT-H & 57.12 & 54.30\tnote{*} \\
          & SAM (Box) \cite{kirillov2023segment} & arxiv23 & ViT-H & 89.35 & 81.09\tnote{*} \\
          & SAM 2-Image (1 Point) \cite{ravi2024sam} & arxiv24 & ViT-H & 77.14 & 73.76\tnote{*} \\
          & SAM 2-Image (Box) \cite{ravi2024sam} & arxiv24 & ViT-H & \textbf{90.18} & \textbf{81.97}\tnote{*} \\
          & SAM 2-Video (1 Point) \cite{ravi2024sam} & arxiv24 & ViT-H & 65.19 & 57.59\tnote{*} \\
    \midrule
    \multirow{3}[0]{*}{Prompt-free} 
          & DINOv3-S \cite{dinov3} & arxiv25 & ViT-S & 86.05 & 39.86 \\
          & DINOv3-B \cite{dinov3} & arxiv25 & ViT-B & 89.04 & 46.37 \\
          & DINOv3-L \cite{dinov3} & arxiv25 & ViT-L & \textbf{92.19} & 63.97 \\
    \bottomrule
    \end{tabular*}%
    }
    \footnotesize
  \end{threeparttable}
  \label{tab:overall_res_endovis18}%
\end{table}

\begin{table*}[ht]
\centering
\resizebox{\textwidth}{!}{
\begin{tabular}{l|cccccc|ccc}
\hline
\multirow{2}{*}{\textbf{Methods}} &
\multicolumn{6}{c|}{\textbf{Dice Metrics} $\uparrow$} &
\multirow{2}{*}{\textbf{mIOU} $\uparrow$} &
\multirow{2}{*}{\textbf{Prec.} $\uparrow$} &
\multirow{2}{*}{\textbf{Recall} $\uparrow$} \\
\cline{2-7}
& Avg. & NDBE & CA & HGD & polyp & Susp. & & & \\
\hline
TransUnet\cite{chen2021}  & 82.45 & 83.85 & 83.53 & 83.31 & 82.54 & 79.00 & 76.53 & 82.57 & 73.65 \\
Unet \cite{ronneberger2015u}      & 64.84 & 69.62 & 59.99 & 69.89 & 74.27 & 50.41 & 51.58 & 59.23 & 61.70 \\
HarDNet \cite{huang2021hardnetmsegsimpleencoderdecoderpolyp}    & 85.31 & 88.15 & 85.78 & 89.13 & 84.51 & 78.94 & 79.66 & 88.05 & 82.24 \\
Swin UNETR \cite{hatamizadeh2022}& 77.40 & 78.89 & 71.86 & 82.12 & 79.48 & 74.63 & 70.64 & 74.88 & 71.05 \\
ESFPNet \cite{Chang_2023}   & 76.58 & 76.71 & 77.19 & 80.52 & 77.67 & 70.82 & 67.56 & 74.19 & 72.72 \\
DUAT  \cite{tang2022duatdualaggregationtransformernetwork}      & 84.89 & 87.91 & \textbf{88.61} & 87.59 & 84.62 & 75.70 & 79.07 & 86.27 & 77.20 \\
FCBFormer \cite{sanderson2022fcn} & 82.74 & 58.90 & 85.12 & 84.91 & 81.12 & 76.67 & 78.09 & 86.65 & 75.73 \\
MSRF-Net \cite{srivastava2022msrfnetmultiscaleresidualfusion}  & 83.42 & 84.08 & 82.56 & 87.77 & 84.84 & 77.86 & 77.51 & 84.79 & 76.86 \\
GMSRF-Net \cite{srivastava2021gmsrfnetimprovedgeneralizabilityglobal} & 82.88 & 84.86 & 83.29 & 83.57 & 84.48 & 78.19 & 76.73 & 82.13 & 79.28 \\
Polyp-PVT \cite{dong2023PolypPVT} & 84.45 & 86.87 & 85.25 & 88.90 & 83.51 & 77.73 & 78.09 & 86.09 & 77.18 \\
PNS+ \cite{Ji_2022}      & 75.52 & 75.36 & 75.62 & 79.37 & 77.71 & 69.54 & 66.23 & 72.84 & 72.20 \\

EAT \cite{11184616} & \textbf{88.02} & \textbf{91.89} & 86.80 & \textbf{92.51} & 86.34 & \textbf{82.57} & \textbf{84.85} & \textbf{91.42} & \textbf{79.45} \\
\hline
\textbf{DINOv3-S} \cite{dinov3}& 71.50 & 86.14 & 77.58 & 74.59 & \textbf{93.93} & 25.25 & 60.30 & 67.73 & 57.47 \\
\textbf{DINOv3-B} \cite{dinov3}& 73.56 & 84.92 & 77.07 & 73.23 & 88.82 & 43.79 & 60.43 & 66.54 & 56.84 \\
\textbf{DINOv3-L} \cite{dinov3}& 72.90 & 88.69 & 78.60 & 82.16 & 92.08 & 22.96 & 62.49 & 72.27 & 59.73 \\
\hline
\end{tabular}
}
\caption{Quantitative comparison of DINOv3 with other SOTA methods on disease segmentation on EDD 2020 dataset. Results for other SOTA methods are derived from \cite{11184616}.}
\label{tab:edd}
\end{table*}

\subsection{3D Classification Results}
\noindent\textbf{Classification on 3D CT images:} 
\label{exp:ctrate}
For 3D classification on the CT-RATE~\cite{ct-rate} dataset, DINOv3 establishes a powerful new baseline, significantly outperforming prior models. As detailed in Table~\ref{tab:CT_RATE}, all DINOv3 variants, using either k-NN or linear probing, achieve substantially higher scores across all metrics compared to the CT-Net and CT-CLIP~\cite{ctclip} baselines. While this comparison is favourable, it is worth noting that CT-CLIP was pre-trained on only 50k samples, unlike other large-scale models such as BiomedCLIP~\cite{BiomedCLIP} which used 15M samples, making a direct comparison of foundation model pre-training scale complex. Notably, DINOv3-B with linear probing achieves an AUC of 0.798, a considerable improvement over CT-CLIP's 0.731. This strong performance demonstrates that DINOv3's 2D features, when aggregated slice-wise, are highly effective for volumetric CT classification tasks without requiring any medical-specific pre-training.

\begin{table}[ht!]
\centering
\caption{3D classification results on the CT-RATE \cite{ct-rate} dataset, evaluated across 18 clinical categories (e.g., \textit{Medical material}, \textit{Arterial wall calcification}, \textit{Cardiomegaly}). The top block shows baseline performance from CT-Net and CT-CLIP. The bottom block evaluates DINOv3 backbones using two methods: a k-NN classifier on frozen features (left) and a trained linear probing (right). For each method, the best result per metric is in \textbf{bold} and the second-best is \underline{underlined}.}
\begin{tabular}{lcccccccc}
\toprule
\textbf{Methods} & \textbf{AUC} & \textbf{ACC} & \textbf{Precision} & \textbf{Recall} & \textbf{AUC} & \textbf{ACC} & \textbf{Precision} & \textbf{Recall} \\
\midrule
CLIP \cite{ctclip} & \multicolumn{4}{c}{\textbf{CT-Net}~\cite{MLAbnormalityCT}} & \multicolumn{4}{c}{\textbf{CT-CLIP}} \\
\cmidrule(lr){2-5} \cmidrule(lr){6-9}
      & 0.629 & 0.657 & 0.263 & \textbf{0.575} & 0.731 & 0.707 & 0.323 & 0.663 \\
\midrule
      & \multicolumn{4}{c}{\textbf{k-NN}} & \multicolumn{4}{c}{\textbf{Linear Probing}} \\
\cmidrule(lr){2-5} \cmidrule(lr){6-9}
DINOv3-S \cite{dinov3} & \underline{0.716} & \underline{0.791} & 0.350 & 0.275 & 0.778 & \underline{0.722} & 0.370 & \underline{0.690} \\
DINOv3-B \cite{dinov3}  & \textbf{0.737} & 0.729 & \underline{0.374} & \underline{0.541} & \textbf{0.798} & \textbf{0.741} & \textbf{0.390} & 0.688 \\
DINOv3-L \cite{dinov3} & 0.709 & \textbf{0.797} & \textbf{0.423} & 0.250 & \underline{0.791} & \underline{0.722} & \underline{0.374} & \textbf{0.728} \\
\bottomrule
\end{tabular}
\label{tab:CT_RATE}
\end{table}

\subsection{3D Segmentation Results}

\noindent\textbf{Segmentation on MSD benchmarks:}
\label{exp:msd}
On the diverse MSD benchmark, DINOv3 shows mixed and generally modest performance compared to state-of-the-art segmentation-specific models like nnU-Net \cite{isensee2021}, as shown in Table \ref{tab:msd_results_part1} and \ref{tab:msd_results_part2}. While DINOv3-L achieves the best Dice scores on a few tasks (e.g., Lung, and Spleen), its overall average performance lags behind top transformer-based and classic methods. This suggests that although its features provide a reasonable starting point, the simple frozen-backbone, slice-by-slice approach is insufficient to compete with fully optimized 3D segmentation architectures. More advanced adapters may therefore be required to effectively translate strong 2D visual features into 3D dense prediction tasks.

\begin{figure}[ht!]
\centering
\includegraphics[width=\textwidth]{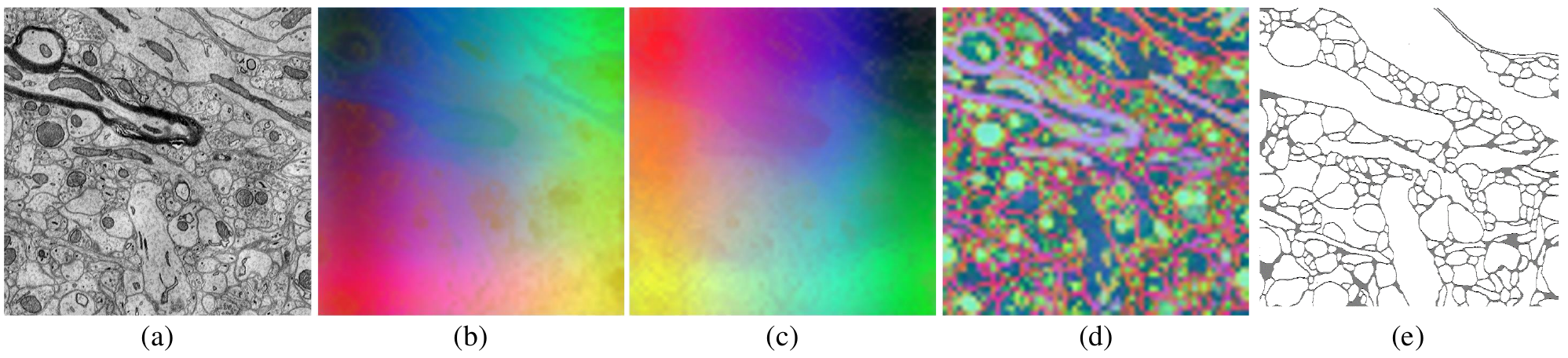}
\caption{Visualization of a slice from the AC3/4 \cite{AC34} dataset and feature embeddings. (a) Raw EM image. (b–d) Feature embeddings extracted from DINOv3-S/16 (b), DINOv3-B/16 (c), and DINOv3-L/16 (d) models, visualized by projecting the first three principal components into RGB space. (e) Corresponding affinity map derived from the raw image.}
\label{fig:em_visual}
\end{figure}

\noindent\textbf{Neuron Segmentation on EM images:} 
\label{exp:em}
DINOv3's features fail catastrophically on EM neuron segmentation. As shown in Tables \ref{tab:cremi} and \ref{tab:AC34_wafer4} , for both CREMI \cite{CREMI2016} and AC3/4 \cite{AC34} datasets, the error rates (VOI and ARAND, where lower is better) for all DINOv3 models are an order of magnitude worse than classic segmentation methods. The visualizations in Figure \ref{fig:em_visual} suggest that the features learned from natural images are too coarse and lack the high-frequency textural detail necessary to delineate the intricate and complex boundaries of neurons in EM volumes. This represents a clear limitation where the domain shift from natural images to EM is too significant for the features to be useful.

\begin{figure}[ht!]
\centering
\includegraphics[width=0.9\textwidth]{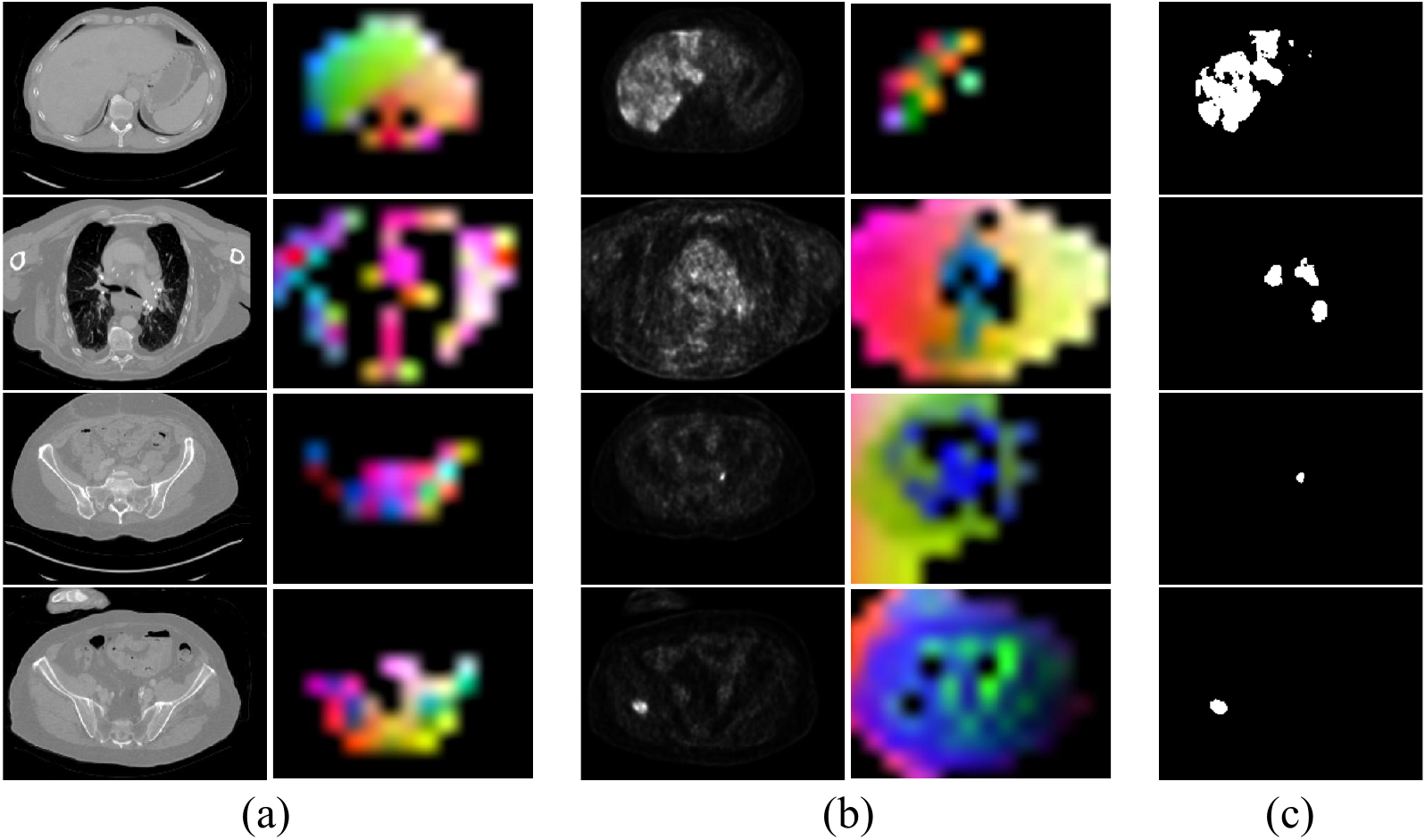}
    \caption{Visualization of the first three principal components derived from PCA on image patches. (a) CT images and (b) PET images are shown with their respective PCA visualizations, where each of the first three components is mapped to a color channel. (c) The resulting tumor region can be isolated by thresholding the first principal component to remove the background.}
\label{fig:pet_ct_visualization}
\end{figure}

\noindent\textbf{Tumor segmentation on FDG-PET/CT images:} 
\label{exp:pet}
Similar to its performance on EM images, DINOv3 performs very poorly on tumor segmentation in PET/CT scans across both the AutoPET-II~\cite{autopetii} and HECKTOR 2022~\cite{hecktor2022} datasets. As shown in Table~\ref{tab:autopet_hecktor_all_methods}, its segmentation performance is drastically lower than established models. This failure likely highlights DINOv3's inability to interpret PET data, as its self-supervised visual features are primarily attuned to anatomical structure. This hypothesis is supported by the visualizations in Figure~\ref{fig:pet_ct_visualization}, which suggest that while DINOv3 features capture anatomical shapes in CT images, they fail to isolate the metabolically active tumor regions in PET images, still focusing on underlying structural patterns. Ultimately, the functional information in PET imaging represents a fundamental departure from the structural patterns in natural images, creating a domain shift that DINOv3's pre-trained features cannot overcome.

\begin{table}[ht!]
\centering
\begin{tabular}{l|c|c|c|c|c}
\hline
\textbf{Methods} & \textbf{Task01} & \textbf{Task02} & \textbf{Task03} & \textbf{Task04} & \textbf{Task05} \\
& \textbf{(Brain)} & \textbf{(Heart)} & \textbf{(Liver)} & \textbf{(Hippo.)} & \textbf{(Prostate)} \\
\hline
\multicolumn{6}{c}{\textit{Supervised Learning Methods}} \\
\hline
3D U-Net \cite{cicek2016} & 72.4 & 81.3 & 91.2 & 76.8 & 82.1 \\
V-Net \cite{milletari2016} & 71.8 & 83.7 & 90.8 & 78.2 & 84.3 \\
nnU-Net \cite{isensee2021} & \textbf{78.9} & \textbf{89.4} & \textbf{96.2} & \textbf{84.1} & \textbf{91.3} \\
TransUNet \cite{chen2021} & 74.2 & 85.1 & 93.4 & 79.6 & 86.7 \\
SwinUNETR \cite{hatamizadeh2022} & \underline{76.5} & \underline{87.3} & \underline{94.7} & \underline{81.2} & \underline{88.9} \\
UNETR \cite{hatamizadeh2021} & 75.1 & 86.2 & 93.9 & 80.4 & 87.5 \\
\hline
\multicolumn{6}{c}{\textit{Self-Supervised Methods (Linear Fine-tuning)}} \\
\hline
MAE-ViT-B/16 \cite{he2022} & 62.1 & 73.8 & 82.3 & 68.4 & 75.2 \\
MAE-ViT-L/16 \cite{he2022} & 64.5 & 76.2 & \textbf{84.1} & 71.2 & 78.1 \\
SimCLR \cite{chen2020} & 58.9 & 70.1 & 79.8 & 64.7 & 72.5 \\
MoCo-v3 \cite{chen2021moco} & 61.3 & 73.2 & 81.6 & 67.9 & 74.8 \\
SwAV \cite{caron2020} & 60.2 & 71.8 & 80.4 & 66.1 & 73.6 \\
BYOL \cite{grill2020} & 60.8 & 72.5 & 80.9 & 67.3 & 74.1 \\
DINOv3-S \cite{dinov3} & 65.2 & 77.1 & \underline{83.8} & 72.6 & 78.9 \\
DINOv3-B \cite{dinov3} & \textbf{66.8} & \textbf{78.2} & \textbf{84.1} & \textbf{75.3} & \underline{79.8} \\
DINOv3-L \cite{dinov3} & \underline{65.9} & \underline{77.6} & 83.5 & \underline{73.8} & \textbf{80.5} \\
\hline
\end{tabular}
\caption{3D segmentation Dice scores (\%) across Medical Segmentation Decathlon (MSD) benchmark tasks (Part 1: Tasks 01-05). For each method, the best result per metric is in \textbf{bold} and the second-best is \underline{underlined}.}
\label{tab:msd_results_part1}
\end{table}

\FloatBarrier

\begin{table}[ht!]
\centering
\begin{tabular}{l|c|c|c|c|c|c}
\hline
\textbf{Methods} & \textbf{Task06} & \textbf{Task07} & \textbf{Task08} & \textbf{Task09} & \textbf{Task10} & \textbf{Average} \\
& \textbf{(Lung)} & \textbf{(Pancreas)} & \textbf{(Hepatic)} & \textbf{(Spleen)} & \textbf{(Colon)} & \\
\hline
\multicolumn{7}{c}{\textit{Supervised Learning Methods}} \\
\hline
3D U-Net \cite{unet3d} & 67.9 & 71.5 & 55.3 & 87.6 & 42.1 & 72.8 \\
V-Net \cite{vnet} & 66.4 & 73.2 & 57.1 & 89.2 & 41.8 & 73.7 \\
nnU-Net \cite{isensee2021} & \textbf{75.8} & \textbf{82.7} & \textbf{67.9} & \textbf{94.8} & \textbf{52.6} & \textbf{81.4} \\
TransUNet \cite{chen2021} & 70.3 & 76.8 & 59.4 & 91.2 & 45.7 & 76.2 \\
SwinUNETR \cite{hatamizadeh2022} & \underline{72.6} & \underline{78.9} & \underline{62.1} & \underline{92.8} & \underline{47.3} & \underline{78.2} \\
UNETR \cite{hatamizadeh2021} & 71.8 & 77.4 & 60.7 & 91.9 & 46.2 & 77.1 \\
\hline
\multicolumn{7}{c}{\textit{Self-Supervised Methods (Linear Fine-tuning)}} \\
\hline
MAE-ViT-B/16 \cite{he2022} & 61.4 & 66.8 & 48.9 & 81.2 & 35.4 & 65.6 \\
MAE-ViT-L/16 \cite{he2022} & 64.1 & 69.3 & 52.1 & 84.8 & 38.7 & 68.3 \\
SimCLR \cite{chen2020} & 57.8 & 63.2 & 45.1 & 77.4 & 31.9 & 62.1 \\
MoCo-v3 \cite{chen2021moco} & 60.9 & 66.1 & 48.2 & 80.6 & 35.1 & 64.8 \\
SwAV \cite{caron2020} & 59.1 & 64.7 & 46.8 & 78.9 & 33.2 & 63.5 \\
BYOL \cite{grill2020} & 60.2 & 65.4 & 47.5 & 79.7 & 34.6 & 64.2 \\
DINOv3-S \cite{dinov3} & 65.8 & 70.2 & 53.4 & 82.9 & 40.1 & 69.0 \\
DINOv3-B \cite{dinov3} & \textbf{73.1} & \textbf{78.9} & \textbf{64.8} & \underline{86.4} & \textbf{49.1} & \textbf{71.3} \\
DINOv3-L \cite{dinov3} & \underline{72.4} & \underline{78.2} & \underline{63.7} & \textbf{91.2} & \underline{47.8} & \underline{71.0} \\
\hline
\end{tabular}
\caption{3D segmentation Dice scores (\%) across Medical Segmentation Decathlon (MSD) benchmark tasks (Part 2: Tasks 06-10). For each method, the best result per metric is in \textbf{bold} and the second-best is \underline{underlined}.}
\label{tab:msd_results_part2}
\end{table}

\begin{table}[ht!]
\centering
\caption{Quantitative comparison of different methods on the CREMI datasets. For each metric, the best result is in \textbf{bold} and the second-best is \underline{underlined}. Note that all reported metrics are lower-is-better.}
{
\fontsize{9}{11}\selectfont
\setlength{\tabcolsep}{1mm} 
\definecolor{Gray}{gray}{0.88}
\begin{tabular}{l|cccc|cccc|cccc}
\toprule
\multirow{2}{*}{Method} & \multicolumn{4}{c|}{\textbf{CREMI-A}} & \multicolumn{4}{c|}{\textbf{CREMI-B}} & \multicolumn{4}{c}{\textbf{CREMI-C}} \\
\cmidrule{2-5} \cmidrule{6-9} \cmidrule{10-13}
& $VOI_s$ & $VOI_m$ & $VOI$ & $ARAND$ & $VOI_s$ & $VOI_m$ & $VOI$ & $ARAND$ & $VOI_s$ & $VOI_m$ & $VOI$ & $ARAND$ \\
\midrule
\multicolumn{13}{c}{\textit{Classic Segmentation Methods}} \\
\midrule
Superhuman~\cite{lee2017superhuman} & 0.399 & 0.241 & 0.640 & 0.089 & 0.554 & \underline{0.222} & \underline{0.776} & 0.048 & 0.820 & 0.338 & 1.158 & 0.179 \\
MALA~\cite{MALA} & 0.398 & \underline{0.236} & 0.634 & 0.085 & 0.589 & 0.261 & 0.850 & 0.041 & 0.842 & 0.332 & 1.174 & 0.162 \\
PEA~\cite{huangwei} & \underline{0.329} & 0.298 & 0.626 & 0.091 & \underline{0.411} & 0.374 & 0.785 & 0.041 & \underline{0.745} & 0.446 & 1.191 & 0.169 \\
APViT~\cite{APViT} & 0.445 & 0.260 & 0.704 & 0.117 & 0.579 & \textbf{0.201} & 0.781 & \underline{0.032} & 0.884 & \underline{0.234} & 1.118 & \textbf{0.110} \\
LSD~\cite{lsd} & 0.393 & \textbf{0.217} & \underline{0.610} & \textbf{0.070} & 0.538 & 0.267 & 0.805 & 0.122 & 0.836 & \textbf{0.230} & \underline{1.065} & 0.150 \\
CAD~\cite{CAD} & \textbf{0.313} & 0.252 & \textbf{0.565} & \underline{0.079} & \textbf{0.379} & 0.305 & \textbf{0.684} & \textbf{0.030} & \textbf{0.738} & 0.322 & \textbf{1.060} & \underline{0.149} \\
\hline
\multicolumn{13}{c}{\textit{DINOv3 Foundation Models (Linear Probing)}} \\
\hline
DINOv3-S~\cite{dinov3} & 2.147 & 1.795 & 3.942 & 0.642
 & 3.048 & 3.660 & 6.708 & 0.543
 & 3.890 & 5.257 & 9.147 & 0.917
 \\
DINOv3-B~\cite{dinov3} & 1.849 & 1.693 & 3.542 & 0.611
 & 2.535 & 3.256 & 5.791 & 0.506
 & 3.457 & 4.089 & 7.546 & 0.795
 \\
DINOv3-L~\cite{dinov3} & 0.793 & 0.991 & 1.784 & 0.448
 & 1.852 & 1.417 & 3.269 & 0.235
 & 2.557 & 1.836 & 4.393 & 0.461 \\
\bottomrule
\end{tabular}
\label{tab:cremi}
}
\end{table}

\begin{table}[ht!]
\centering
\caption{Quantitative comparison of different methods on AC3/4 and Wafer4 datasets. We compare classic segmentation methods and DINOv3 foundation models (linear probing). Best results are in \textbf{bold}, and second best are \underline{underlined}. Note: Reported metrics are all lower-is-better.}
{
\fontsize{9}{11}\selectfont
\definecolor{Gray}{gray}{0.88}
\begin{tabular}{l|cccc|cccc}
\toprule
\multirow{2}{*}{Method} & \multicolumn{4}{c|}{\textbf{AC3/4}} & \multicolumn{4}{c}{\textbf{Wafer4}} \\
\cmidrule{2-5}
\cmidrule{6-9}
& $VOI_s$ & $VOI_m$ & $VOI$ & $ARAND$ & $VOI_s$ & $VOI_m$ & $VOI$ & $ARAND$ \\
\hline
\multicolumn{9}{c}{\textit{Classic Segmentation Methods}} \\
\hline
Superhuman~\cite{lee2017superhuman}  & 0.597 & 0.433 & 1.031 & 0.179 & 0.452 & 0.166 & 0.618 & 0.041 \\
MALA~\cite{MALA} & 0.677 & 0.457 & 1.134 & 0.166 & 0.455 & 0.158 & 0.613 & 0.036 \\
PEA~\cite{huangwei} & \underline{0.552} & 0.498 & 1.050 & 0.209 & \underline{0.421} & 0.172 & 0.593 & 0.034 \\
APViT~\cite{APViT} & 0.767 & \textbf{0.204} & 0.976 & \textbf{0.078} & 0.581 & \underline{0.123} & 0.704 & 0.036 \\
LSD~\cite{lsd} & 0.633 & \underline{0.280} & \underline{0.913} & 0.093 & 0.445 & \textbf{0.115} & \underline{0.560} & \textbf{0.026} \\
CAD~\cite{CAD} & \textbf{0.533} & 0.351 & \textbf{0.884} & \underline{0.081} & \textbf{0.415} & 0.144 & \textbf{0.559} & \underline{0.030} \\
\hline
\multicolumn{9}{c}{\textit{DINOv3 Foundation Models (Linear Probing)}} \\
\hline
DINOv3-S~\cite{dinov3} & 3.813 & 5.252 & 8.965 & 0.825
 & 4.298 & 2.705 & 7.003 & 0.331 \\
DINOv3-B~\cite{dinov3} & 3.070 & 2.009 & 5.079 & 0.274
 & 3.564 & 1.722 & 5.286 & 0.189 \\
DINOv3-L~\cite{dinov3} & 1.821 & 0.950 & 2.771 & 0.268
 & 2.061 & 0.568 & 2.629 & 0.115 \\
\bottomrule
\end{tabular}
\label{tab:AC34_wafer4}
}
\end{table}

\begin{table*}[ht!]
    \centering
    \caption{Performance comparison of different methods on AutoPET-II and HECKTOR 2022 datasets across CT and PET modalities. Best results across all methods are in \textbf{bold} and second best are \underline{underlined}. Notably, HD95 is NaN, which means the prediction is all background.}
    \small
    \setlength{\tabcolsep}{3pt}
    \renewcommand{\arraystretch}{1.2}
    \begin{tabular}{l c cccc cccc}
    \hline
    \multirow{2}{*}{\centering\textbf{Methods}} & \multirow{2}{*}{\textbf{Modality}} & \multicolumn{4}{c}{\textbf{AutoPET-II}} & \multicolumn{4}{c}{\textbf{HECKTOR 2022}} \\
    \cmidrule(lr){3-6} \cmidrule(lr){7-10}
    & & \textbf{Dice} & \textbf{HD95} & \textbf{Prec.} & \textbf{Rec.} & \textbf{Dice} & \textbf{HD95} & \textbf{Prec.} & \textbf{Rec.} \\
    \hline
    \multicolumn{10}{c}{\textit{Classic Segmentation Methods}} \\
    \hline
    UNet \cite{unet3d} & CT+PET & 59.41 & 241.31 & 62.32 & 70.74 & 50.25 & 65.03 & 72.13 & 41.50 \\
    VNet \cite{vnet} & CT+PET & 53.21 & 242.78 & 53.21 & 60.85 & \underline{55.61} & 41.46 & \underline{78.21} & 46.01 \\
    UNETR \cite{hatamizadeh2021} & CT+PET & 51.49 & 257.30 & 51.49 & 61.03 & 48.10 & 73.27 & 70.71 & 39.11 \\
    Swin UNETR \cite{swinunetr} & CT+PET & \underline{62.24} & 242.07 & \underline{62.91} & 73.30 & 44.56 & 103.02 & 62.43 & 37.55 \\
    VSmTrans \cite{vsmtrans} & CT+PET & \textbf{62.46} & 223.88 & \textbf{65.19} & 70.92 & 52.91 & 78.03 & 61.91 & \textbf{50.97} \\
    UNETR++ \cite{unetr++} & CT+PET & 36.50 & 178.57 & 36.50 & 60.16 & 29.95 & 27.74 & 61.84 & 21.75 \\
    U-KAN \cite{u_kan} & CT+PET & 60.67 & \textbf{70.91} & 62.03 & 72.94 & \textbf{55.89} & \textbf{23.48} & 77.72 & \underline{46.89} \\
    \hline
    \multicolumn{10}{c}{\textit{Multimodal Segmentation Methods}} \\ \hline
    Nestedformer \cite{nestedformer} & CT+PET & 61.38 & 265.51 & 61.38 & 64.29 & 40.17 & 72.95 & 63.22 & 32.59 \\
    A2FSeg \cite{a2fseg} & CT+PET & 60.86 & \underline{131.48} & 60.86 & \textbf{76.10} & 40.90 & 32.95 & 77.02 & 30.57 \\
    H-DenseFormer \cite{hdenseformer} & CT+PET & 61.50 & 252.98 & 61.41 & \underline{75.76} & 46.79 & 34.84 & \textbf{78.33} & 35.31 \\
    \hline
    \multicolumn{10}{c}{\textit{DINOv3 Foundation Models (Linear Probing)}} \\
    \hline
    DINOv3-S/16 \cite{dinov3} & CT & 0.00 & 25475.80 & 0.00 & 0.00 & 0.00 & NaN & 0.00 & 0.00 \\
    DINOv3-B/16 \cite{dinov3} & CT & 0.00 & 21394.57 & 0.00 & 0.00 & 0.00 & 7541.56 & 0.03 & 0.00 \\
    DINOv3-L/16 \cite{dinov3} & CT & 0.00 & 11637.64 & 0.39 & 0.00 & 0.00 & NaN & 0.00 & 0.00 \\
    \hline
    DINOv3-S/16 \cite{dinov3} & PET & 7.10 & 13940.53 & 4.37 & 48.14 & 6.44 & 10641.95 & 5.92 & 17.37 \\
    DINOv3-B/16 \cite{dinov3} & PET & 8.74 & 14114.07 & 5.43 & 54.38 & 21.41 & 7919.81 & 37.03 & 20.57 \\
    DINOv3-L/16 \cite{dinov3} & PET & 10.87 & 13611.39 & 6.85 & 64.96 & 9.43 & 10329.33 & 9.40 & 25.93 \\
    \hline
    DINOv3-S/16 \cite{dinov3} & CT+PET & 9.06 & 13456.42 & 5.32 & 65.87 & 40.13 & 4294.74 & 52.82 & 40.67 \\
    DINOv3-B/16 \cite{dinov3} & CT+PET & 14.53 & 13188.93 & 9.50 & 49.06 & 39.50 & 5032.80 & 45.37 & 45.18 \\
    DINOv3-L/16 \cite{dinov3} & CT+PET & 12.17 & 13418.89 & 7.50 & 71.16 & 30.86 & 8808.90 & 34.99 & 39.98 \\
    \hline
    \end{tabular}
    \label{tab:autopet_hecktor_all_methods}
\end{table*}

\subsection{2D and 3D Registration}

\begin{figure}[h!]
    \centering
    \includegraphics[width=0.8\linewidth]{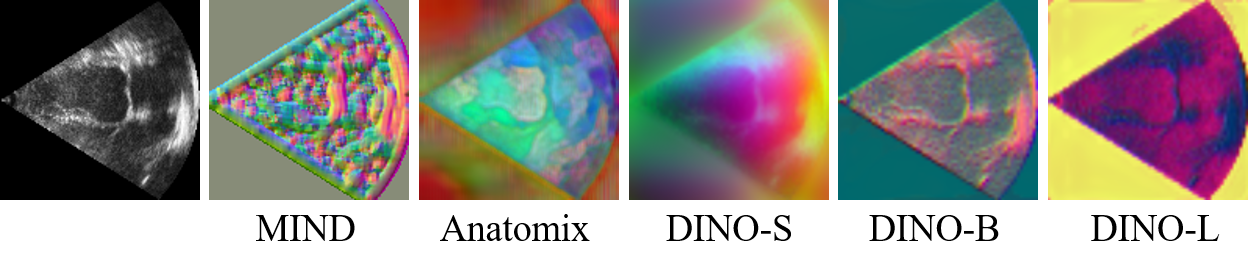}
    \caption{Visualization of ultrasound features extracted by different encoders. Each image displays the first three principal components of the feature representation, mapped to RGB color channels.}
    \label{fig:CardiacFeat}
\end{figure}

\noindent\textbf{2D Registration of Cardiac Ultrasound Images:}
For 2D cardiac image registration on the CAMUS dataset~\cite{leclerc2019camus}, DINOv3, along with other feature-based registration methods, does not match the performance of VoxelMorph. Table~\ref{camus} summarizes the quantitative results. Notably, MIND features fail drastically in this scenario; as shown in Figure~\ref{fig:CardiacFeat}, the feature maps are corrupted and lack meaningful anatomical information. Consequently, we do not concatenate MIND features with DINOv3 features for this comparison. Anatomix exhibits similar limitations, showing slightly better anatomical detail but noticeable dispersion outside the cone region. DINO-S features also show spatial dispersion, although the myocardial structure begins to emerge. In contrast, DINO-B and DINO-L produce better anatomical representations. However, the highlighted regions are not confined solely to the myocardium, as noise is also emphasized. These observations underscore the need for ultrasound-specific foundation models that explicitly account for the unique characteristics of ultrasound imaging.

\begin{table}[h!]
    \centering
    \caption{Quantitative results of 2D registration on CAMUS. The results are presented as mean ± standard deviation. Best results across all methods are in \textbf{bold}.}
    \label{camus}
    \setlength{\tabcolsep}{8pt} 
    \begin{tabular}{l ccc}
        \toprule
        Methods & Dice $\uparrow$ & HD $\downarrow$ & ASD $\downarrow$ \\
        \midrule
        Unregistered & 0.7397 ± 0.1101 & 7.0401 ± 2.1989 & 3.3067 ± 1.2462 \\
        VoxelMorph~\cite{balakrishnan2019voxelmorph} & \textbf{0.8592 ± 0.0500} & \textbf{4.6211 ± 1.9272} & \textbf{1.8961 ± 0.7095} \\
        MIND+ConvexAdam~\cite{siebert2021cvxAdam} & 0.7100 ± 0.1261 & 8.5663 ± 3.0280 & 3.6472 ± 1.3983 \\
        Anatomix~\cite{dey2024anatomix}+ConvexAdam & 0.8012 ± 0.0902 & 5.5769 ± 1.7707 & 2.5564 ± 0.9419 \\
        \midrule
        \multicolumn{4}{c}{\textit{DINOv3 Foundation Models (Zero-Shot)}} \\
        \midrule
        DINO-S+ConvexAdam & 0.8401 ± 0.0895 & 5.4652 ± 2.9305 & 2.1236 ± 1.0167 \\
        DINO-B+ConvexAdam & 0.8315 ± 0.0927 & 5.7703 ± 2.9907 & 2.2565 ± 1.0976 \\
        DINO-L+ConvexAdam & 0.8431 ± 0.0886 & 5.4670 ± 3.0653 & 2.1061 ± 1.0291 \\
        \bottomrule
    \end{tabular}
\end{table}

\begin{figure}[h!]
    \centering
    \includegraphics[width=\linewidth]{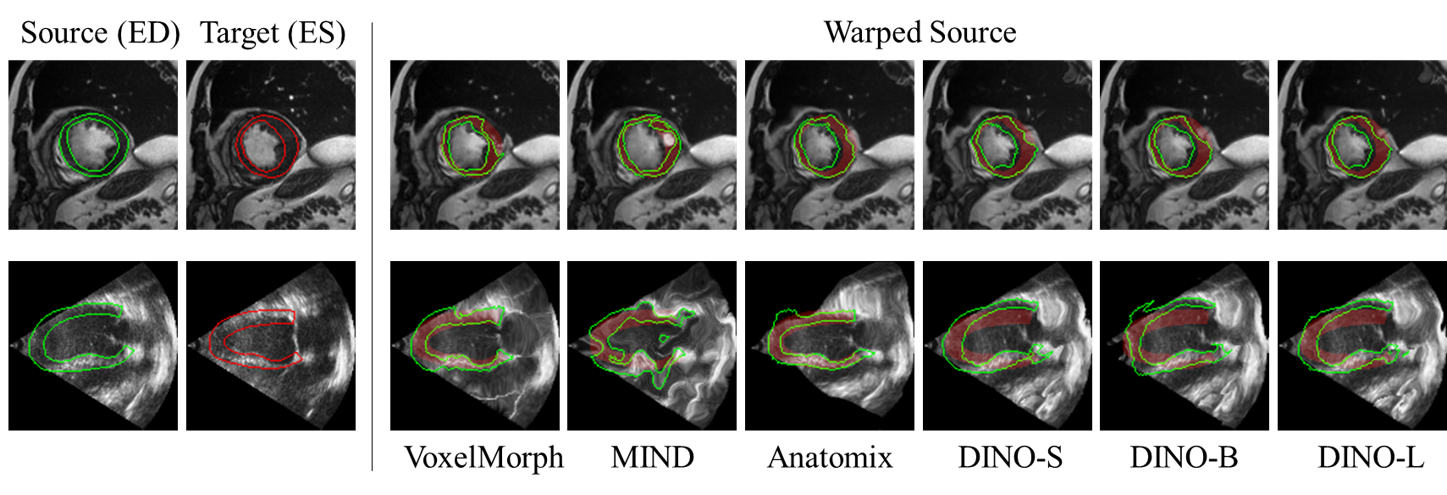}
    \caption{Qualitative results of cardiac image registration on the ACDC (top row) and CAMUS (second row) datasets. The first two columns show source and target images with ground truth myocardium segmentation contours. The subsequent columns display warped source images produced by different registration methods. Each warped image is overlaid with warped myocardium contours, with red highlighting the ground truth End-Systole (ES) myocardium.}
    \label{fig:CardiacReg}
\end{figure}

\noindent\textbf{3D Registration of Cardiac MRI Volumes:} 
For 3D image registration on the ACDC dataset~\cite{ACDC}, DINOv3 establishes a strong baseline, moderately outperforming other methods. Table~\ref{acdc} presents the results for cardiac MRI volume registration. Quantitatively, feature-based registration methods are on par with the widely used VoxelMorph method. However, qualitative differences are evident in the warped segmentation maps. As illustrated in the first row of Figure~\ref{fig:CardiacReg}, DINOv3 demonstrates superior correspondence, particularly in scenarios involving occlusions. In instances where tissues inside the myocardial segmentation ring exhibit intensity differences, DINOv3 features yield the smoothest and most well-aligned results.

\begin{table}[h!]
    \centering
    \caption{Quantitative results of 3D registration on ACDC. The results are presented as mean ± standard deviation. Best results across all methods are in \textbf{bold}.}
    \label{acdc}
    \setlength{\tabcolsep}{6pt} 
    \begin{tabular}{l ccc}
        \toprule
        Methods & Dice $\uparrow$ & HD (mm) $\downarrow$ & ASD (mm) $\downarrow$ \\
        \midrule
        Unregistered & 0.5889 ± 0.1706 & 11.1254 ± 4.0871 & 5.0708 ± 2.6632 \\
        VoxelMorph~\cite{balakrishnan2019voxelmorph} & 0.7383 ± 0.1220 & \textbf{8.3389 ± 4.3419} & 3.4038 ± 2.0344 \\
        MIND+ConvexAdam~\cite{siebert2021cvxAdam} & 0.7499 ± 0.1168 & 8.6275 ± 4.8917 & 3.4302 ± 2.1990 \\
        Anatomix~\cite{dey2024anatomix}+ConvexAdam & 0.7566 ± 0.1132 & 8.4479 ± 4.8564 & 3.3477 ± 2.1367 \\
        \midrule
        \multicolumn{4}{c}{\textit{DINOv3 Foundation Models (Zero-Shot)}} \\
        \midrule
        DINO-S+MIND+ConvexAdam & 0.7480 ± 0.1173 & 8.5931 ± 4.8235 & 3.4153 ± 2.1625 \\
        DINO-B+MIND+ConvexAdam & \textbf{0.7593 ± 0.1124} & 8.3439 ± 4.8711 & \textbf{3.2957 ± 2.1244} \\
        DINO-L+MIND+ConvexAdam & 0.7481 ± 0.1171 & 8.5652 ± 4.7769 & 3.4101 ± 2.1583 \\
        \bottomrule
    \end{tabular}
\end{table}

\section{Findings}
\begin{tcolorbox}[colback=mygold, colframe=mygold!75!black, boxrule=0pt, arc=2mm, left=1mm, right=1mm, top=1mm, bottom=1mm]
\begin{center}
\textit{\textbf{F1: DINOv3's natural-image features excel on some medical tasks but fail on modalities with a large domain shift.}}
\end{center}
\end{tcolorbox}
DINOv3 \cite{dinov3}, pretrained solely on natural images, establishes a strong new baseline in the medical domain without any medical specific pre training. It demonstrates impressive performance, showing results comparable to domain specific models like BiomedCLIP \cite{BiomedCLIP} and CT-CLIP \cite{ctclip}, and sometimes even outperforming them in certain scenarios. Specifically, it achieves comparable performance on 2D chest X ray classification (NIH 14 \cite{Wang2017ChestXRay8HC} and RSNA Pneumonia \cite{rsna2018pneumonia} datasets) and sets a strong new baseline for 3D CT classification (CT-RATE \cite{ct-rate} dataset). In the field of endoscopic imaging, DINOv3 also delivers competitive results; it achieves state-of-the-art performance in binary instrument segmentation on the EndoVis18 dataset, although it does not consistently surpass specialized supervised methods in fine-grained classification tasks. Furthermore, in cardiac MRI registration, DINOv3 features demonstrate superior correspondence compared to standard methods, particularly in scenarios involving occlusions. However, DINOv3 performs poorly on WSI classification, EM, and PET segmentation.

This performance disparity can be hypothesized to stem from the object centric nature of DINOv3 \cite{dinov3} pretraining. Since the model learned from a vast corpus of natural images from Instagram, its visual features are highly attuned to capturing structures and shapes. This explains its success in modalities like X-ray, CT, and endoscopy, where many diagnostic patterns are linked to macroscopic structural abnormalities. In contrast, its performance degrades significantly on image modalities where the visual characteristics differ greatly. For WSI, analysis relies on fine grained textural and cellular patterns, which are less represented in DINOv3 object focused feature space. For EM, the model features lack the high frequency textural detail required to delineate intricate neuronal boundaries. The shift is even more pronounced for PET, as these scans visualize functional metabolic activity, a fundamental departure from the structural patterns in natural images that DINOv3 is primed to recognize.

\begin{tcolorbox}[colback=mygold, colframe=mygold!75!black, boxrule=0pt, arc=2mm, left=1mm, right=1mm, top=1mm, bottom=1mm]
\begin{center}
\textit{\textbf{F2: Scaling laws from natural images do not consistently transfer to the medical domain.}}
\end{center}
\end{tcolorbox}
The report finds that DINOv3 does not consistently follow the expected scaling laws in the medical domain. Contrary to trends in natural image tasks, increasing the model size (e.g., from DINOv3 S to DINOv3 L) or using higher input resolutions does not reliably lead to better performance. For instance, on the NIH 14 chest X ray dataset, performance peaks at a $512\times512$ resolution before declining at higher resolutions. This inconsistent scaling behavior is observed across different tasks and datasets, indicating that larger models are not consistently able to achieve the best performance. This suggests that simply using a larger model or finer features is not a guaranteed strategy for improvement in medical imaging.

\begin{tcolorbox}[colback=mygold, colframe=mygold!75!black, boxrule=0pt, arc=2mm, left=1mm, right=1mm, top=1mm, bottom=1mm]
\begin{center}
\textit{\textbf{F3: The benefits of scaling are not uniformly transferable across diverse medical tasks and modalities.}}
\end{center}
\end{tcolorbox}
The advantages gained from scaling are not uniformly transferable, with different tasks exhibiting markedly different behaviors. This is particularly evident in 2D classification; for both chest X-ray and WSI analysis, larger models can paradoxically underperform smaller ones. In contrast, for 3D CT classification, increasing model scale is generally beneficial, though the improvement is not always monotonic. A third distinct pattern appears in 3D segmentation, where larger DINOv3 models typically outperform their smaller counterparts. Remarkably, on certain Medical Segmentation Decathlon tasks like Lung segmentation, the aggregated 2D features from DINOv3 can achieve performance comparable to the strong nnU-Net baseline. This underscores the potential of leveraging powerful 2D visual priors for complex 3D tasks, indicating that these features are not universal and vary significantly depending on the specific medical task and modality.

\subsection{Limitations of this Report}
While this report presents a comprehensive benchmark across diverse tasks and modalities, it has several limitations. First, our analysis focuses exclusively on the DINOv3 model family and does not include a comparative evaluation against other foundation models~\cite{bolya2025perception}. Second, our experiments are restricted to a linear probing protocol with a frozen backbone; we do not explore the potential benefits of full fine-tuning or parameter-efficient adaptation methods~\cite{hu2022lora, ma2024segment}. Finally, although the selected datasets are diverse, they are not exhaustive. Our benchmark does not cover all medical imaging modalities, such as 4D cardiac MRI \cite{zhang2025towards,zhang2024whole}, or all relevant tasks, such as 3D reconstruction~\cite{jian2025timeflow, bubeck2025latent}.

\section{Conclusion}
\subsection{Summary of Findings}
This report establishes DINOv3 as a strong off the shelf encoder for a range of medical imaging tasks, particularly those with visual characteristics similar to natural images such as CT and X ray analysis. Despite being trained exclusively on non medical data, it sets a strong baseline and can achieve performance comparable to domain specific models in certain scenarios. However, our findings highlight critical limitations: DINOv3 performance deteriorates significantly in domains like WSI, EM, and PET, where there may be even greater shifts between training and target distributions. Furthermore, we observe that the scaling laws that govern performance on natural images do not consistently apply in the medical domain; larger models and higher resolutions do not reliably yield better results, revealing complex and task dependent scaling behaviors.

\subsection{Future Directions}
Based on our findings, several promising research avenues emerge. First, to bridge the performance gap in specialized domains, future work should move beyond linear probing and investigate parameter efficient fine tuning methods to adapt DINOv3 features for new domains. Second, for volumetric tasks, there is a clear need to develop more sophisticated 2D to 3D adapters that can more effectively translate the powerful slice wise features for dense 3D prediction tasks like segmentation. Finally, the high quality of DINOv3 features in modalities like CT could be leveraged for other complex tasks, such as enforcing multi view consistency in 3D reconstruction from 2D slices or improving medical image registration.

{
\small
\bibliographystyle{IEEEtran}
\bibliography{paper}

\begin{thebibliography}{100}
\providecommand{\url}[1]{#1}
\csname url@samestyle\endcsname
\providecommand{\newblock}{\relax}
\providecommand{\bibinfo}[2]{#2}
\providecommand{\BIBentrySTDinterwordspacing}{\spaceskip=0pt\relax}
\providecommand{\BIBentryALTinterwordstretchfactor}{4}
\providecommand{\BIBentryALTinterwordspacing}{\spaceskip=\fontdimen2\font plus
\BIBentryALTinterwordstretchfactor\fontdimen3\font minus \fontdimen4\font\relax}
\providecommand{\BIBforeignlanguage}[2]{{%
\expandafter\ifx\csname l@#1\endcsname\relax
\typeout{** WARNING: IEEEtran.bst: No hyphenation pattern has been}%
\typeout{** loaded for the language `#1'. Using the pattern for}%
\typeout{** the default language instead.}%
\else
\language=\csname l@#1\endcsname
\fi
#2}}
\providecommand{\BIBdecl}{\relax}
\BIBdecl

\bibitem{OpenAI2022ChatGPT}
\BIBentryALTinterwordspacing
OpenAI, ``Chatgpt,'' 2022. [Online]. Available: \url{https://openai.com/blog/chatgpt}
\BIBentrySTDinterwordspacing

\bibitem{kaplan2020scaling}
J.~Kaplan, S.~McCandlish, T.~Henighan, T.~B. Brown, B.~Chess, R.~Child, S.~Gray, A.~Radford, J.~Wu, and D.~Amodei, ``Scaling laws for neural language models,'' \emph{arXiv preprint arXiv:2001.08361}, 2020.

\bibitem{alabdulmohsin2022revisiting}
I.~M. Alabdulmohsin, B.~Neyshabur, and X.~Zhai, ``Revisiting neural scaling laws in language and vision,'' \emph{Advances in Neural Information Processing Systems}, vol.~35, pp. 22\,300--22\,312, 2022.

\bibitem{xie2023data}
Z.~Xie, Z.~Zhang, Y.~Cao, Y.~Lin, Y.~Wei, Q.~Dai, and H.~Hu, ``On data scaling in masked image modeling,'' in \emph{Proceedings of the IEEE/CVF Conference on Computer Vision and Pattern Recognition}, 2023, pp. 10\,365--10\,374.

\bibitem{el2024scalable}
A.~El-Nouby, M.~Klein, S.~Zhai, M.~A. Bautista, A.~Toshev, V.~Shankar, J.~M. Susskind, and A.~Joulin, ``Scalable pre-training of large autoregressive image models,'' \emph{arXiv preprint arXiv:2401.08541}, 2024.

\bibitem{pan2025beyond}
J.~Pan, B.~Jian, P.~Hager, Y.~Zhang, C.~Liu, F.~Jungmann, H.~B. Li, C.~You, J.~Wu, J.~Zhu \emph{et~al.}, ``Beyond benchmarks: Dynamic, automatic and systematic red-teaming agents for trustworthy medical language models,'' \emph{arXiv preprint arXiv:2508.00923}, 2025.

\bibitem{pan2025medvlm}
J.~Pan, C.~Liu, J.~Wu, F.~Liu, J.~Zhu, H.~B. Li, C.~Chen, C.~Ouyang, and D.~Rueckert, ``Medvlm-r1: Incentivizing medical reasoning capability of vision-language models (vlms) via reinforcement learning,'' \emph{arXiv preprint arXiv:2502.19634}, 2025.

\bibitem{fan2025scaling}
D.~Fan, S.~Tong, J.~Zhu, K.~Sinha, Z.~Liu, X.~Chen, M.~Rabbat, N.~Ballas, Y.~LeCun, A.~Bar \emph{et~al.}, ``Scaling language-free visual representation learning,'' \emph{arXiv preprint arXiv:2504.01017}, 2025.

\bibitem{dinov2}
M.~Oquab, T.~Darcet, T.~Moutakanni, H.~Vo, M.~Szafraniec, V.~Khalidov, P.~Fernandez, D.~Haziza, F.~Massa, A.~El-Nouby \emph{et~al.}, ``Dinov2: Learning robust visual features without supervision,'' \emph{arXiv preprint arXiv:2304.07193}, 2023.

\bibitem{caron2021emerging}
M.~Caron, H.~Touvron, I.~Misra, H.~J{\'e}gou, J.~Mairal, P.~Bojanowski, and A.~Joulin, ``Emerging properties in self-supervised vision transformers,'' in \emph{Proceedings of ICCV}, 2021, pp. 9650--9660.

\bibitem{dinov3}
O.~Sim{\'e}oni, H.~V. Vo, M.~Seitzer, F.~Baldassarre, M.~Oquab, C.~Jose, V.~Khalidov, M.~Szafraniec, S.~Yi, M.~Ramamonjisoa \emph{et~al.}, ``Dinov3,'' \emph{arXiv preprint arXiv:2508.10104}, 2025.

\bibitem{yang2025segdino}
S.~Yang, H.~Wang, Z.~Xing, S.~Chen, and L.~Zhu, ``Segdino: An efficient design for medical and natural image segmentation with dino-v3,'' \emph{arXiv preprint arXiv:2509.00833}, 2025.

\bibitem{li2025meddinov3}
Y.~Li, Y.~Wu, Y.~Lai, M.~Hu, and X.~Yang, ``Meddinov3: How to adapt vision foundation models for medical image segmentation?'' \emph{arXiv preprint arXiv:2509.02379}, 2025.

\bibitem{Wang2017ChestXRay8HC}
X.~Wang, Y.~Peng, L.~Lu, Z.~Lu, M.~Bagheri, and R.~M. Summers, ``Chestx-ray8: Hospital-scale chest x-ray database and benchmarks on weakly-supervised classification and localization of common thorax diseases,'' in \emph{Proceedings of CVPR}, 2017, pp. 3462--3471.

\bibitem{CONCH}
M.~Y. Lu, B.~Chen, D.~F. Williamson, R.~J. Chen, I.~Liang, T.~Ding, G.~Jaume, I.~Odintsov, L.~P. Le, G.~Gerber \emph{et~al.}, ``A visual-language foundation model for computational pathology,'' \emph{Nature medicine}, vol.~30, no.~3, pp. 863--874, 2024.

\bibitem{ct-rate}
I.~E. Hamamci, S.~Er, C.~Wang, F.~Almas, A.~G. Simsek, S.~N. Esirgun, I.~Doga, O.~F. Durugol, W.~Dai, M.~Xu \emph{et~al.}, ``Developing generalist foundation models from a multimodal dataset for 3d computed tomography,'' \emph{arXiv preprint arXiv:2403.17834}, 2024.

\bibitem{BiomedCLIP}
S.~Zhang, Y.~Xu, N.~Usuyama, J.~Bagga, R.~Tinn, S.~Preston, R.~Rao, M.~Wei, N.~Valluri, C.~Wong \emph{et~al.}, ``Large-scale domain-specific pretraining for biomedical vision-language processing,'' \emph{arXiv preprint arXiv:2303.00915}, 2023.

\bibitem{rsna2018pneumonia}
\BIBentryALTinterwordspacing
A.~Stein, C.~Wu, C.~Carr, G.~Shih, J.~Dulkowski, kalpathy, L.~Chen, L.~Prevedello, M.~Kohli, M.~McDonald, Peter, P.~Culliton, S.~Halabi, and T.~Xia, ``{RSNA} pneumonia detection challenge,'' \url{https://www.kaggle.com/competitions/rsna-pneumonia-detection-challenge}, 2018. [Online]. Available: \url{https://www.kaggle.com/competitions/rsna-pneumonia-detection-challenge}
\BIBentrySTDinterwordspacing

\bibitem{mgca}
F.~Wang, Y.~Zhou, S.~Wang, V.~Vardhanabhuti, and L.~Yu, ``Multi-granularity cross-modal alignment for generalized medical visual representation learning,'' \emph{Advances in neural information processing systems}, vol.~35, pp. 33\,536--33\,549, 2022.

\bibitem{Camelyon16}
B.~E. Bejnordi, M.~Veta, P.~J. Van~Diest, B.~Van~Ginneken, N.~Karssemeijer, G.~Litjens, J.~A. Van Der~Laak, M.~Hermsen, Q.~F. Manson, M.~Balkenhol \emph{et~al.}, ``Diagnostic assessment of deep learning algorithms for detection of lymph node metastases in women with breast cancer,'' \emph{Jama}, vol. 318, no.~22, pp. 2199--2210, 2017.

\bibitem{Camelyon17}
P.~Bandi, O.~Geessink, Q.~Manson, M.~Van~Dijk, M.~Balkenhol, M.~Hermsen, B.~E. Bejnordi, B.~Lee, K.~Paeng, A.~Zhong \emph{et~al.}, ``From detection of individual metastases to classification of lymph node status at the patient level: the camelyon17 challenge,'' \emph{IEEE transactions on medical imaging}, vol.~38, no.~2, pp. 550--560, 2018.

\bibitem{attrimil}
L.~Cai, S.~Huang, Y.~Zhang, J.~Lu, and Y.~Zhang, ``Attrimil: Revisiting attention-based multiple instance learning for whole-slide pathological image classification from a perspective of instance attributes,'' \emph{Medical Image Analysis}, p. 103631, 2025.

\bibitem{BCNB}
F.~Xu, C.~Zhu, W.~Tang, Y.~Wang, Y.~Zhang, J.~Li, H.~Jiang, Z.~Shi, J.~Liu, and M.~Jin, ``Predicting axillary lymph node metastasis in early breast cancer using deep learning on primary tumor biopsy slides,'' \emph{Frontiers in oncology}, vol.~11, p. 759007, 2021.

\bibitem{CLAM}
M.~Y. Lu, D.~F. Williamson, T.~Y. Chen, R.~J. Chen, M.~Barbieri, and F.~Mahmood, ``Data-efficient and weakly supervised computational pathology on whole-slide images,'' \emph{Nature biomedical engineering}, vol.~5, no.~6, pp. 555--570, 2021.

\bibitem{Smedsrud2021}
P.~H. Smedsrud, V.~Thambawita, S.~A. Hicks, H.~Gjestang, O.~O. Nedrejord, E.~N{\ae}ss, H.~Borgli, D.~Jha, T.~J.~D. Berstad, S.~L. Eskeland, M.~Lux, H.~Espeland, A.~Petlund, D.~T.~D. Nguyen, E.~Garcia-Ceja, D.~Johansen, P.~T. Schmidt, E.~Toth, H.~L. Hammer, T.~de~Lange, M.~A. Riegler, and P.~Halvorsen, ``{Kvasir-Capsule, a video capsule endoscopy dataset},'' \emph{Scientific Data}, vol.~8, no.~1, p. 142, 2021.

\bibitem{Wang2022}
Z.~Wang, B.~Lu, Y.~Long, F.~Zhong, T.~H. Cheung, Q.~Dou, and Y.~Liu, ``Autolaparo: {A} new dataset of integrated multi-tasks for image-guided surgical automation in laparoscopic hysterectomy,'' \emph{CoRR}, vol. abs/2208.02049, 2022.

\bibitem{leclerc2019camus}
S.~Leclerc, E.~Smistad, J.~Pedrosa, A.~{\O}stvik, F.~Cervenansky, F.~Espinosa, T.~Espeland, E.~A.~R. Berg, P.-M. Jodoin, T.~Grenier \emph{et~al.}, ``Deep learning for segmentation using an open large-scale dataset in 2d echocardiography,'' \emph{IEEE transactions on medical imaging}, vol.~38, no.~9, pp. 2198--2210, 2019.

\bibitem{EndoVis2018}
\BIBentryALTinterwordspacing
M.~Allan, S.~Kondo, S.~Bodenstedt, S.~Leger, R.~Kadkhodamohammadi, I.~Luengo, F.~Fuentes{-}Hurtado, E.~Flouty, A.~K. Mohammed, M.~Pedersen, A.~Kori, A.~Varghese, G.~Krishnamurthi, D.~Rauber, R.~Mendel, C.~Palm, S.~Bano, G.~Saibro, C.~Shih, H.~Chiang, J.~Zhuang, J.~Yang, V.~Iglovikov, A.~Dobrenkii, M.~Reddiboina, A.~Reddy, X.~Liu, C.~Gao, M.~Unberath, M.~Azizian, D.~Stoyanov, L.~Maier{-}Hein, and S.~Speidel, ``2018 robotic scene segmentation challenge,'' \emph{CoRR}, vol. abs/2001.11190, 2020. [Online]. Available: \url{https://arxiv.org/abs/2001.11190}
\BIBentrySTDinterwordspacing

\bibitem{gonzalez2020isinet}
C.~Gonz{\'a}lez, L.~Bravo-S{\'a}nchez, and P.~Arbelaez, ``Isinet: an instance-based approach for surgical instrument segmentation,'' in \emph{Medical Image Computing and Computer Assisted Intervention--MICCAI 2020: 23rd International Conference, Lima, Peru, October 4--8, 2020, Proceedings, Part III 23}.\hskip 1em plus 0.5em minus 0.4em\relax Springer, 2020, pp. 595--605.

\bibitem{https://doi.org/10.21227/f8xg-wb80}
\BIBentryALTinterwordspacing
S.~Ali, B.~Braden, D.~Lamarque, S.~Realdon, A.~Bailey, R.~Cannizzaro, N.~Ghatwary, J.~Rittscher, C.~Daul, and J.~East, ``Endoscopy disease detection and segmentation (edd2020),'' 2020. [Online]. Available: \url{https://ieee-dataport.org/competitions/endoscopy-disease-detection-and-segmentation-edd2020}
\BIBentrySTDinterwordspacing

\bibitem{muller2025medical}
G.~M{\"u}ller-Franzes, F.~Khader, R.~Siepmann, T.~Han, J.~N. Kather, S.~Nebelung, and D.~Truhn, ``Medical slice transformer for improved diagnosis and explainability on 3d medical images with dinov2,'' \emph{Scientific Reports}, vol.~15, no.~1, p. 23979, 2025.

\bibitem{msd}
M.~Antonelli, A.~Reinke, S.~Bakas, K.~Farahani, A.~Kopp-Schneider, B.~A. Landman, G.~Litjens, B.~Menze, O.~Ronneberger, R.~M. Summers \emph{et~al.}, ``The medical segmentation decathlon,'' \emph{Nature communications}, vol.~13, no.~1, p. 4128, 2022.

\bibitem{wu2024voco}
L.~Wu, J.~Zhuang, and H.~Chen, ``Voco: A simple-yet-effective volume contrastive learning framework for 3d medical image analysis,'' in \emph{Proceedings of the IEEE/CVF conference on computer vision and pattern recognition}, 2024, pp. 22\,873--22\,882.

\bibitem{CREMI2016}
CREMI, ``Miccai challenge on circuit reconstruction from electron microscopy images,'' \url{https://cremi.org/}, 2016.

\bibitem{AC34}
N.~Kasthuri, K.~J. Hayworth, D.~R. Berger, R.~L. Schalek, J.~A. Conchello, S.~Knowles-Barley, D.~Lee, A.~V{\'a}zquez-Reina, V.~Kaynig, T.~R. Jones \emph{et~al.}, ``Saturated reconstruction of a volume of neocortex,'' \emph{Cell}, vol. 162, no.~3, pp. 648--661, 2015.

\bibitem{autopetii}
K.~T. Gatidis~S, ``A whole-body fdg-pet/ct dataset with manually annotated tumor lesions (fdg-pet-ct-lesions),'' \emph{The Cancer Imaging Archive}, vol. 226, 2022.

\bibitem{hecktor2022}
V.~Oreiller, V.~Andrearczyk, M.~Jreige, S.~Boughdad, H.~Elhalawani, J.~Castelli, M.~Vallieres, S.~Zhu, J.~Xie, Y.~Peng \emph{et~al.}, ``Head and neck tumor segmentation in pet/ct: the hecktor challenge,'' \emph{Medical image analysis}, vol.~77, p. 102336, 2022.

\bibitem{ACDC}
O.~Bernard, A.~Lalande, C.~Zotti, F.~Cervenansky, X.~Yang, P.-A. Heng, I.~Cetin, K.~Lekadir, O.~Camara, M.~A. Gonzalez~Ballester, G.~Sanroma, S.~Napel, S.~Petersen, G.~Tziritas, E.~Grinias, M.~Khened, V.~A. Kollerathu, G.~Krishnamurthi, M.-M. Rohé, X.~Pennec, M.~Sermesant, F.~Isensee, P.~Jäger, K.~H. Maier-Hein, P.~M. Full, I.~Wolf, S.~Engelhardt, C.~F. Baumgartner, L.~M. Koch, J.~M. Wolterink, I.~Išgum, Y.~Jang, Y.~Hong, J.~Patravali, S.~Jain, O.~Humbert, and P.-M. Jodoin, ``Deep learning techniques for automatic mri cardiac multi-structures segmentation and diagnosis: Is the problem solved?'' \emph{IEEE Transactions on Medical Imaging}, vol.~37, no.~11, pp. 2514--2525, 2018.

\bibitem{ilse2018attention}
M.~Ilse, J.~Tomczak, and M.~Welling, ``Attention-based deep multiple instance learning,'' in \emph{International conference on machine learning}.\hskip 1em plus 0.5em minus 0.4em\relax PMLR, 2018, pp. 2127--2136.

\bibitem{MALA}
J.~Funke, F.~Tschopp, W.~Grisaitis, A.~Sheridan, C.~Singh, S.~Saalfeld, and S.~C. Turaga, ``Large scale image segmentation with structured loss based deep learning for connectome reconstruction,'' \emph{IEEE transactions on pattern analysis and machine intelligence}, vol.~41, no.~7, pp. 1669--1680, 2018.

\bibitem{song2024dino}
X.~Song, X.~Xu, and P.~Yan, ``Dino-reg: General purpose image encoder for training-free multi-modal deformable medical image registration,'' in \emph{International Conference on Medical Image Computing and Computer-Assisted Intervention}.\hskip 1em plus 0.5em minus 0.4em\relax Springer, 2024, pp. 608--617.

\bibitem{voi}
J.~Nunez-Iglesias, R.~Kennedy, T.~Parag, J.~Shi, and D.~B. Chklovskii, ``Machine learning of hierarchical clustering to segment 2d and 3d images,'' \emph{PloS one}, vol.~8, no.~8, p. e71715, 2013.

\bibitem{arand}
I.~Arganda-Carreras, S.~C. Turaga, D.~R. Berger, D.~Cire{\c{s}}an, A.~Giusti, L.~M. Gambardella, J.~Schmidhuber, D.~Laptev, S.~Dwivedi, J.~M. Buhmann \emph{et~al.}, ``Crowdsourcing the creation of image segmentation algorithms for connectomics,'' \emph{Frontiers in neuroanatomy}, vol.~9, p. 152591, 2015.

\bibitem{UNI}
R.~J. Chen, T.~Ding, M.~Y. Lu, D.~F. Williamson, G.~Jaume, A.~H. Song, B.~Chen, A.~Zhang, D.~Shao, M.~Shaban \emph{et~al.}, ``Towards a general-purpose foundation model for computational pathology,'' \emph{Nature medicine}, vol.~30, no.~3, pp. 850--862, 2024.

\bibitem{ResNet}
K.~He, X.~Zhang, S.~Ren, and J.~Sun, ``Deep residual learning for image recognition,'' in \emph{Proceedings of the IEEE conference on computer vision and pattern recognition}, 2016, pp. 770--778.

\bibitem{JOSEPH2025131325}
\BIBentryALTinterwordspacing
J.~Joseph, S.~N. George, and K.~Raja, ``Vapcaps: A novel variance-based attention network with imbalance aware loss for better pathology detection in video capsule endoscopy,'' \emph{Neurocomputing}, vol. 655, p. 131325, 2025. [Online]. Available: \url{https://www.sciencedirect.com/science/article/pii/S0925231225019976}
\BIBentrySTDinterwordspacing

\bibitem{Li2026STSANet}
\BIBentryALTinterwordspacing
Y.~Li, G.~Zhao, C.~Li, W.~Shi, Z.~Jiang, Z.~Zhang, and G.~Feng, ``Stsanet: Spatial temporal-self-aggregation network for surgical phase recognition,'' \emph{Information Fusion}, vol. 126, p. 103646, 2026. [Online]. Available: \url{https://www.sciencedirect.com/science/article/pii/S1566253525007183}
\BIBentrySTDinterwordspacing

\bibitem{GMSRFNet}
A.~Srivastava, S.~Chanda, D.~Jha, U.~Pal, and S.~Ali, ``Gmsrf-net: An improved generalizability with global multi-scale residual fusion network for polyp segmentation,'' in \emph{2022 26th International Conference on Pattern Recognition (ICPR)}, 2022, pp. 4321--4327.

\bibitem{ConvMix153620}
A.~Trockman and J.~Kolter, ``Patches are all you need?'' \emph{Transactions on Machine Learning Research}, 2022.

\bibitem{ConViTS}
S.~d'Ascoli, H.~Touvron, M.~Leavitt, A.~Morcos, G.~Biroli, and L.~Sagun, ``Convit: Improving vision transformers with soft convolutional inductive biases,'' \emph{Journal of Statistical Mechanics: Theory and Experiment}, vol. 2022, no.~11, p. 114005, 2022.

\bibitem{SwinS}
X.~Dong, J.~Bao, D.~Chen, W.~Zhang, N.~Yu, L.~Yuan, D.~Chen, and B.~Guo, ``Cswin transformer: A general vision transformer backbone with cross-shaped windows,'' in \emph{2022 IEEE/CVF Conference on Computer Vision and Pattern Recognition (CVPR)}.\hskip 1em plus 0.5em minus 0.4em\relax IEEE Computer Society, 2022, pp. 12\,114--12\,124.

\bibitem{FocalConvNet}
A.~Srivastava, N.~Tomar, and D.~Jha, ``Video capsule endoscopy classification using focal modulation guided convolutional neural network,'' in \emph{Proceedings. IEEE International Symposium on Computer-Based Medical Systems}, vol. 2022, 2022, pp. 323--328.

\bibitem{VatsEtAl}
A.~Vats, M.~Pedersen, A.~Mohammed, and {\O}.~Hovde., ``Learning more for free - a multi task learning approach for improved pathology classification in capsule endoscopy,'' in \emph{Medical Image Computing and Computer Assisted Intervention – MICCAI 2021}, M.~de~Bruijne, P.~Cattin, S.~Cotin, N.~Padoy, S.~Speidel, Y.~Zheng, and C.~Essert, Eds.\hskip 1em plus 0.5em minus 0.4em\relax Cham: Springer International Publishing, 2021, pp. 3--13.

\bibitem{APINet}
O.~Yet, T.~Rassem, M.~Rahman, and M.~Rahman, ``Improved attentive pairwise interaction (api-net) for fine-grained image classification,'' in \emph{2021 Emerging Technology in Computing, Communication and Electronics (ETCCE)}, 2021, pp. 1--6.

\bibitem{Jin2018SVRCNet}
Y.~Jin, Q.~Dou, H.~Chen, L.~Yu, J.~Qin, C.-W. Fu, and P.-A. Heng, ``Sv-rcnet: Workflow recognition from surgical videos using recurrent convolutional network,'' \emph{IEEE Transactions on Medical Imaging}, vol.~37, no.~5, pp. 1114--1126, 2018.

\bibitem{Jin2021TMRNet}
Y.~Jin, Y.~Long, C.~Chen, Z.~Zhao, Q.~Dou, and P.-A. Heng, ``Temporal memory relation network for workflow recognition from surgical video,'' \emph{IEEE Transactions on Medical Imaging}, vol.~40, no.~7, pp. 1911--1923, 2021.

\bibitem{Gao2021}
X.~Gao, Y.~Jin, Y.~Long, Q.~Dou, and P.~Heng, ``Trans-svnet: Accurate phase recognition from surgical videos via hybrid embedding aggregation transformer,'' in \emph{Medical Image Computing and Computer Assisted Intervention - {MICCAI} 2021 - 24th International Conference, Strasbourg, France, September 27 - October 1, 2021, Proceedings, Part {IV}}, ser. Lecture Notes in Computer Science, M.~de~Bruijne, P.~C. Cattin, S.~Cotin, N.~Padoy, S.~Speidel, Y.~Zheng, and C.~Essert, Eds., vol. 12904.\hskip 1em plus 0.5em minus 0.4em\relax Springer, 2021, pp. 593--603.

\bibitem{Liu2025LoViT}
Y.~Liu, M.~Boels, L.~Garcia-Peraza-Herrera, T.~Vercauteren, P.~Dasgupta, A.~Granados, and S.~Ourselin, ``Lovit: Long video transformer for surgical phase recognition,'' \emph{Medical Image Analysis}, vol.~99, p. 103366, 2025.

\bibitem{yu2024sam2roboticsurgery}
\BIBentryALTinterwordspacing
J.~Yu, A.~Wang, W.~Dong, M.~Xu, M.~Islam, J.~Wang, L.~Bai, and H.~Ren, ``Sam 2 in robotic surgery: An empirical evaluation for robustness and generalization in surgical video segmentation,'' 2024. [Online]. Available: \url{https://arxiv.org/abs/2408.04593}
\BIBentrySTDinterwordspacing

\bibitem{ronneberger2015u}
O.~Ronneberger, P.~Fischer, and T.~Brox, ``U-net: Convolutional networks for biomedical image segmentation,'' in \emph{Medical Image Computing and Computer-Assisted Intervention--MICCAI 2015: 18th International Conference, Munich, Germany, October 5-9, 2015, Proceedings, Part III 18}.\hskip 1em plus 0.5em minus 0.4em\relax Springer, 2015, pp. 234--241.

\bibitem{shvets2018automatic}
A.~A. Shvets, A.~Rakhlin, A.~A. Kalinin, and V.~I. Iglovikov, ``Automatic instrument segmentation in robot-assisted surgery using deep learning,'' in \emph{2018 17th IEEE International Conference on Machine Learning and Applications (ICMLA)}, 2018, pp. 624--628.

\bibitem{jin2019incorporating}
Y.~Jin, K.~Cheng, Q.~Dou, and P.-A. Heng, ``Incorporating temporal prior from motion flow for instrument segmentation in minimally invasive surgery video,'' in \emph{Medical Image Computing and Computer Assisted Intervention--MICCAI 2019: 22nd International Conference, Shenzhen, China, October 13--17, 2019, Proceedings, Part V 22}.\hskip 1em plus 0.5em minus 0.4em\relax Springer, 2019, pp. 440--448.

\bibitem{wang2022rethinking}
A.~Wang, M.~Islam, M.~Xu, and H.~Ren, ``Rethinking surgical instrument segmentation: A background image can be all you need,'' in \emph{International Conference on Medical Image Computing and Computer-Assisted Intervention}.\hskip 1em plus 0.5em minus 0.4em\relax Springer, 2022, pp. 355--364.

\bibitem{islam2021st}
M.~Islam, V.~Vibashan, C.~M. Lim, and H.~Ren, ``St-mtl: Spatio-temporal multitask learning model to predict scanpath while tracking instruments in robotic surgery,'' \emph{Medical Image Analysis}, vol.~67, p. 101837, 2021.

\bibitem{islam2020ap}
M.~Islam, V.~Vibashan, and H.~Ren, ``Ap-mtl: Attention pruned multi-task learning model for real-time instrument detection and segmentation in robot-assisted surgery,'' in \emph{2020 IEEE international conference on robotics and automation (ICRA)}.\hskip 1em plus 0.5em minus 0.4em\relax IEEE, 2020, pp. 8433--8439.

\bibitem{seenivasan2022global}
L.~Seenivasan, S.~Mitheran, M.~Islam, and H.~Ren, ``Global-reasoned multi-task learning model for surgical scene understanding,'' \emph{IEEE Robotics and Automation Letters}, vol.~7, no.~2, pp. 3858--3865, 2022.

\bibitem{zhao2022trasetr}
Z.~Zhao, Y.~Jin, and P.-A. Heng, ``Trasetr: track-to-segment transformer with contrastive query for instance-level instrument segmentation in robotic surgery,'' in \emph{2022 International Conference on Robotics and Automation (ICRA)}.\hskip 1em plus 0.5em minus 0.4em\relax IEEE, 2022, pp. 11\,186--11\,193.

\bibitem{baby2023forks}
B.~Baby, D.~Thapar, M.~Chasmai, T.~Banerjee, K.~Dargan, A.~Suri, S.~Banerjee, and C.~Arora, ``From forks to forceps: A new framework for instance segmentation of surgical instruments,'' in \emph{Proceedings of the IEEE/CVF Winter Conference on Applications of Computer Vision}, 2023, pp. 6191--6201.

\bibitem{kirillov2023segment}
A.~Kirillov, E.~Mintun, N.~Ravi, H.~Mao, C.~Rolland, L.~Gustafson, T.~Xiao, S.~Whitehead, A.~C. Berg, W.-Y. Lo \emph{et~al.}, ``Segment anything,'' \emph{arXiv preprint arXiv:2304.02643}, 2023.

\bibitem{ravi2024sam}
N.~Ravi, V.~Gabeur, Y.-T. Hu, R.~Hu, C.~Ryali, T.~Ma, H.~Khedr, R.~R{\"a}dle, C.~Rolland, L.~Gustafson \emph{et~al.}, ``Sam 2: Segment anything in images and videos,'' \emph{arXiv preprint arXiv:2408.00714}, 2024.

\bibitem{chen2021}
J.~Chen, Y.~Lu, Q.~Yu, X.~Luo, E.~Adeli, Y.~Wang, L.~Lu, A.~L. Yuille, and Y.~Zhou, ``Transunet: Transformers make strong encoders for medical image segmentation,'' \emph{arXiv preprint arXiv:2102.04306}, 2021.

\bibitem{huang2021hardnetmsegsimpleencoderdecoderpolyp}
\BIBentryALTinterwordspacing
C.-H. Huang, H.-Y. Wu, and Y.-L. Lin, ``Hardnet-mseg: A simple encoder-decoder polyp segmentation neural network that achieves over 0.9 mean dice and 86 fps,'' 2021. [Online]. Available: \url{https://arxiv.org/abs/2101.07172}
\BIBentrySTDinterwordspacing

\bibitem{hatamizadeh2022}
A.~Hatamizadeh, V.~Nath, Y.~Tang, D.~Yang, H.~R. Roth, and D.~Xu, ``Swin unetr: Swin transformers for semantic segmentation of brain tumors in mri images,'' in \emph{International MICCAI brainlesion workshop}.\hskip 1em plus 0.5em minus 0.4em\relax Springer, 2022, pp. 272--284.

\bibitem{Chang_2023}
\BIBentryALTinterwordspacing
Q.~Chang, D.~Ahmad, J.~Toth, R.~Bascom, and W.~E. Higgins, ``Esfpnet: efficient deep learning architecture for real-time lesion segmentation in autofluorescence bronchoscopic video,'' in \emph{Medical Imaging 2023: Biomedical Applications in Molecular, Structural, and Functional Imaging}, B.~S. Gimi and A.~Krol, Eds.\hskip 1em plus 0.5em minus 0.4em\relax SPIE, Apr. 2023. [Online]. Available: \url{http://dx.doi.org/10.1117/12.2647897}
\BIBentrySTDinterwordspacing

\bibitem{tang2022duatdualaggregationtransformernetwork}
\BIBentryALTinterwordspacing
F.~Tang, Q.~Huang, J.~Wang, X.~Hou, J.~Su, and J.~Liu, ``Duat: Dual-aggregation transformer network for medical image segmentation,'' 2022. [Online]. Available: \url{https://arxiv.org/abs/2212.11677}
\BIBentrySTDinterwordspacing

\bibitem{sanderson2022fcn}
E.~Sanderson and B.~J. Matuszewski, ``Fcn-transformer feature fusion for polyp segmentation,'' in \emph{Annual Conference on Medical Image Understanding and Analysis}.\hskip 1em plus 0.5em minus 0.4em\relax Springer, 2022, pp. 892--907.

\bibitem{srivastava2022msrfnetmultiscaleresidualfusion}
\BIBentryALTinterwordspacing
A.~Srivastava, D.~Jha, S.~Chanda, U.~Pal, H.~D. Johansen, D.~Johansen, M.~A. Riegler, S.~Ali, and P.~Halvorsen, ``Msrf-net: A multi-scale residual fusion network for biomedical image segmentation,'' 2022. [Online]. Available: \url{https://arxiv.org/abs/2105.07451}
\BIBentrySTDinterwordspacing

\bibitem{srivastava2021gmsrfnetimprovedgeneralizabilityglobal}
\BIBentryALTinterwordspacing
A.~Srivastava, S.~Chanda, D.~Jha, U.~Pal, and S.~Ali, ``Gmsrf-net: An improved generalizability with global multi-scale residual fusion network for polyp segmentation,'' 2021. [Online]. Available: \url{https://arxiv.org/abs/2111.10614}
\BIBentrySTDinterwordspacing

\bibitem{dong2023PolypPVT}
D.~Bo, W.~Wenhai, F.~Deng-Ping, L.~Jinpeng, F.~Huazhu, and S.~Ling, ``Polyp-pvt: Polyp segmentation with pyramidvision transformers,'' \emph{CAAI AIR}, 2023.

\bibitem{Ji_2022}
\BIBentryALTinterwordspacing
G.-P. Ji, G.~Xiao, Y.-C. Chou, D.-P. Fan, K.~Zhao, G.~Chen, and L.~Van~Gool, ``Video polyp segmentation: A deep learning perspective,'' \emph{Machine Intelligence Research}, vol.~19, no.~6, p. 531–549, Nov. 2022. [Online]. Available: \url{http://dx.doi.org/10.1007/s11633-022-1371-y}
\BIBentrySTDinterwordspacing

\bibitem{11184616}
Y.~Pang, Y.~Long, Z.~Chen, Y.~Hu, H.~Chen, and Q.~Wang, ``Endoscopic adaptive transformer for enhanced polyp segmentation in endoscopic imaging,'' \emph{IEEE Transactions on Medical Imaging}, pp. 1--1, 2025.

\bibitem{ctclip}
I.~E. Hamamci, S.~Er, C.~Wang, F.~Almas, A.~G. Simsek, S.~N. Esirgun, I.~Doga, O.~F. Durugol, W.~Dai, M.~Xu \emph{et~al.}, ``Developing generalist foundation models from a multimodal dataset for 3d computed tomography,'' \emph{arXiv preprint arXiv:2403.17834}, 2024.

\bibitem{MLAbnormalityCT}
R.~L. Draelos, D.~Dov, M.~A. Mazurowski, J.~Y. Lo, R.~Henao, G.~D. Rubin, and L.~Carin, ``Machine-learning-based multiple abnormality prediction with large-scale chest computed tomography volumes,'' \emph{Medical Image Analysis}, vol.~67, p. 101857, 2021.

\bibitem{isensee2021}
F.~Isensee, P.~F. Jaeger, S.~A. Kohl, J.~Petersen, and K.~H. Maier-Hein, ``nnu-net: a self-configuring method for deep learning-based biomedical image segmentation,'' \emph{Nature methods}, vol.~18, no.~2, pp. 203--211, 2021.

\bibitem{cicek2016}
{\"O}.~{\c{C}}i{\c{c}}ek, A.~Abdulkadir, S.~S. Lienkamp, T.~Brox, and O.~Ronneberger, ``3d u-net: learning dense volumetric segmentation from sparse annotation,'' in \emph{International conference on medical image computing and computer-assisted intervention}.\hskip 1em plus 0.5em minus 0.4em\relax Springer, 2016, pp. 424--432.

\bibitem{milletari2016}
F.~Milletari, N.~Navab, and S.-A. Ahmadi, ``V-net: Fully convolutional neural networks for volumetric medical image segmentation,'' in \emph{2016 fourth international conference on 3D vision (3DV)}.\hskip 1em plus 0.5em minus 0.4em\relax IEEE, 2016, pp. 565--571.

\bibitem{hatamizadeh2021}
A.~Hatamizadeh, Y.~Tang, V.~Nath, D.~Yang, A.~Myronenko, B.~Landman, H.~R. Roth, and D.~Xu, ``Unetr: transformers for 3d medical image segmentation,'' in \emph{Proceedings of the IEEE/CVF winter conference on applications of computer vision}, 2022, pp. 1748--1758.

\bibitem{he2022}
K.~He, X.~Chen, S.~Xie, Y.~Li, P.~Doll{\'a}r, and R.~Girshick, ``Masked autoencoders are scalable vision learners,'' in \emph{Proceedings of the IEEE/CVF conference on computer vision and pattern recognition}, 2022, pp. 16\,000--16\,009.

\bibitem{chen2020}
T.~Chen, S.~Kornblith, M.~Norouzi, and G.~Hinton, ``A simple framework for contrastive learning of visual representations,'' in \emph{International conference on machine learning}.\hskip 1em plus 0.5em minus 0.4em\relax PMLR, 2020, pp. 1597--1607.

\bibitem{chen2021moco}
X.~Chen, S.~Xie, and K.~He, ``An empirical study of training self-supervised vision transformers,'' \emph{arXiv preprint arXiv:2104.02057}, 2021.

\bibitem{caron2020}
M.~Caron, I.~Misra, J.~Mairal, P.~Goyal, P.~Bojanowski, and A.~Joulin, ``Unsupervised learning of visual features by contrasting cluster assignments,'' vol.~33, pp. 9912--9924, 2020.

\bibitem{grill2020}
J.-B. Grill, F.~Strub, F.~Altch{\'e}, C.~Tallec, P.~Richemond, E.~Buchatskaya, C.~Doersch, B.~A. Pires, Z.~Guo, M.~G. Azar \emph{et~al.}, ``Bootstrap your own latent: A new approach to self-supervised learning,'' vol.~33, pp. 21\,271--21\,284, 2020.

\bibitem{unet3d}
{\"O}.~{\c{C}}i{\c{c}}ek, A.~Abdulkadir, S.~S. Lienkamp, T.~Brox, and O.~Ronneberger, ``3d u-net: learning dense volumetric segmentation from sparse annotation,'' in \emph{Medical Image Computing and Computer-Assisted Intervention--MICCAI 2016: 19th International Conference, Athens, Greece, October 17-21, 2016, Proceedings, Part II 19}.\hskip 1em plus 0.5em minus 0.4em\relax Springer, 2016, pp. 424--432.

\bibitem{vnet}
F.~Milletari, N.~Navab, and S.-A. Ahmadi, ``V-net: Fully convolutional neural networks for volumetric medical image segmentation,'' in \emph{2016 fourth international conference on 3D vision (3DV)}.\hskip 1em plus 0.5em minus 0.4em\relax Ieee, 2016, pp. 565--571.

\bibitem{lee2017superhuman}
K.~Lee, J.~Zung, P.~Li, V.~Jain, and H.~S. Seung, ``Superhuman accuracy on the snemi3d connectomics challenge,'' \emph{arXiv preprint arXiv:1706.00120}, 2017.

\bibitem{huangwei}
W.~Huang, S.~Deng, C.~Chen, X.~Fu, and Z.~Xiong, ``Learning to model pixel-embedded affinity for homogeneous instance segmentation,'' in \emph{Proceedings of the AAAI Conference on Artificial Intelligence}, vol.~36, no.~1, 2022, pp. 1007--1015.

\bibitem{APViT}
R.~Sun, N.~Luo, Y.~Pan, H.~Mai, T.~Zhang, Z.~Xiong, and F.~Wu, ``Appearance prompt vision transformer for connectome reconstruction.'' in \emph{IJCAI}, 2023, pp. 1423--1431.

\bibitem{lsd}
A.~Sheridan, T.~M. Nguyen, D.~Deb, W.-C.~A. Lee, S.~Saalfeld, S.~C. Turaga, U.~Manor, and J.~Funke, ``Local shape descriptors for neuron segmentation,'' \emph{Nature methods}, vol.~20, no.~2, pp. 295--303, 2023.

\bibitem{CAD}
X.~Liu, M.~Cai, Y.~Chen, Y.~Zhang, T.~Shi, R.~Zhang, X.~Chen, and Z.~Xiong, ``Cross-dimension affinity distillation for 3d em neuron segmentation,'' in \emph{2024 IEEE/CVF Conference on Computer Vision and Pattern Recognition (CVPR)}.\hskip 1em plus 0.5em minus 0.4em\relax IEEE Computer Society, 2024, pp. 11\,104--11\,113.

\bibitem{swinunetr}
Y.~Tang, D.~Yang, W.~Li, H.~R. Roth, B.~A. Landman, D.~Xu, V.~Nath, and A.~Hatamizadeh, ``Self-supervised pre-training of swin transformers for 3d medical image analysis,'' \emph{Proceedings of CVPR}, pp. 20\,698--20\,708, 2021.

\bibitem{vsmtrans}
T.~Liu, Q.~Bai, D.~A. Torigian, Y.~Tong, and J.~K. Udupa, ``Vsmtrans: A hybrid paradigm integrating self-attention and convolution for 3d medical image segmentation,'' \emph{Medical image analysis}, vol.~98, p. 103295, 2024.

\bibitem{unetr++}
A.~M. Shaker, M.~Maaz, H.~Rasheed, S.~Khan, M.-H. Yang, and F.~S. Khan, ``Unetr++: delving into efficient and accurate 3d medical image segmentation,'' \emph{IEEE Transactions on Medical Imaging}, 2024.

\bibitem{u_kan}
C.~Li \emph{et~al.}, ``U-kan makes strong backbone for medical image segmentation and generation,'' in \emph{Proceedings of the AAAI Conference on Artificial Intelligence}, vol.~39, no.~5, 2025, pp. 4652--4660.

\bibitem{nestedformer}
Z.~Xing, L.~Yu, L.~Wan, T.~Han, and L.~Zhu, ``Nestedformer: Nested modality-aware transformer for brain tumor segmentation,'' in \emph{International conference on medical image computing and computer-assisted intervention}.\hskip 1em plus 0.5em minus 0.4em\relax Springer, 2022, pp. 140--150.

\bibitem{a2fseg}
Z.~Wang and Y.~Hong, ``A2fseg: Adaptive multi-modal fusion network for medical image segmentation,'' in \emph{International Conference on Medical Image Computing and Computer-Assisted Intervention}.\hskip 1em plus 0.5em minus 0.4em\relax Springer, 2023, pp. 673--681.

\bibitem{hdenseformer}
J.~Shi \emph{et~al.}, ``H-denseformer: An efficient hybrid densely connected transformer for multimodal tumor segmentation,'' in \emph{International Conference on Medical Image Computing and Computer-Assisted Intervention}.\hskip 1em plus 0.5em minus 0.4em\relax Springer, 2023, pp. 692--702.

\bibitem{balakrishnan2019voxelmorph}
G.~Balakrishnan, A.~Zhao, M.~R. Sabuncu, J.~Guttag, and A.~V. Dalca, ``Voxelmorph: a learning framework for deformable medical image registration,'' \emph{IEEE transactions on medical imaging}, vol.~38, no.~8, pp. 1788--1800, 2019.

\bibitem{siebert2021cvxAdam}
H.~Siebert, L.~Hansen, and M.~P. Heinrich, ``Fast 3d registration with accurate optimisation and little learning for learn2reg 2021,'' in \emph{International Conference on Medical Image Computing and Computer-Assisted Intervention}.\hskip 1em plus 0.5em minus 0.4em\relax Springer, 2021, pp. 174--179.

\bibitem{dey2024anatomix}
\BIBentryALTinterwordspacing
N.~Dey, B.~Billot, H.~E. Wong, C.~J. Wang, M.~Ren, P.~E. Grant, A.~V. Dalca, and P.~Golland, ``Learning general-purpose biomedical volume representations using randomized synthesis,'' 2024. [Online]. Available: \url{https://arxiv.org/abs/2411.02372}
\BIBentrySTDinterwordspacing

\bibitem{bolya2025perception}
D.~Bolya, P.-Y. Huang, P.~Sun, J.~H. Cho, A.~Madotto, C.~Wei, T.~Ma, J.~Zhi, J.~Rajasegaran, H.~Rasheed \emph{et~al.}, ``Perception encoder: The best visual embeddings are not at the output of the network,'' \emph{arXiv preprint arXiv:2504.13181}, 2025.

\bibitem{hu2022lora}
E.~J. Hu, Y.~Shen, P.~Wallis, Z.~Allen-Zhu, Y.~Li, S.~Wang, L.~Wang, W.~Chen \emph{et~al.}, ``Lora: Low-rank adaptation of large language models.'' \emph{ICLR}, vol.~1, no.~2, p.~3, 2022.

\bibitem{ma2024segment}
J.~Ma, Y.~He, F.~Li, L.~Han, C.~You, and B.~Wang, ``Segment anything in medical images,'' \emph{Nature Communications}, vol.~15, no.~1, p. 654, 2024.

\bibitem{zhang2025towards}
Y.~Zhang, P.~Hager, C.~Liu, S.~Shit, C.~Chen, D.~Rueckert, and J.~Pan, ``Towards cardiac mri foundation models: Comprehensive visual-tabular representations for whole-heart assessment and beyond,'' \emph{arXiv preprint arXiv:2504.13037}, 2025.

\bibitem{zhang2024whole}
Y.~Zhang, C.~Chen, S.~Shit, S.~Starck, D.~Rueckert, and J.~Pan, ``Whole heart 3d+ t representation learning through sparse 2d cardiac mr images,'' in \emph{International Conference on Medical Image Computing and Computer-Assisted Intervention}.\hskip 1em plus 0.5em minus 0.4em\relax Springer, 2024, pp. 359--369.

\bibitem{jian2025timeflow}
B.~Jian, J.~Pan, Y.~Li, F.~Bongratz, R.~Li, D.~Rueckert, B.~Wiestler, and C.~Wachinger, ``Timeflow: Longitudinal brain image registration and aging progression analysis,'' \emph{arXiv preprint arXiv:2501.08667}, 2025.

\bibitem{bubeck2025latent}
N.~Bubeck, S.~Shit, C.~Chen, C.~Zhao, P.~Guo, D.~Yang, G.~Zitzlsberger, D.~Xu, B.~Kainz, D.~Rueckert \emph{et~al.}, ``Latent interpolation learning using diffusion models for cardiac volume reconstruction,'' \emph{arXiv preprint arXiv:2508.13826}, 2025.

\end{thebibliography}
}


\section*{Acknowledgement}
The LaTeX template is built upon Meta’s original template.



\end{document}